\documentclass[twoside,11pt]{article}
\usepackage[preprint]{jmlr2e}
\usepackage[x11names, table]{xcolor}
\usepackage{blindtext}
\usepackage{amsmath}
\newtheorem{observation}{Observation}
\usepackage{tikz}
\usepackage{algorithm}
\usepackage[noend]{algpseudocode}
\usepackage{booktabs}
\graphicspath{{figures/}}
\usepackage{svg}
\usepackage{adjustbox}
\usepackage[newfloat, frozencache, cachedir=.]{minted}
\usepackage{caption}
\newenvironment{code}{\captionsetup{type=listing, format=hang}}{}
\SetupFloatingEnvironment{listing}{name=Code}

\usepackage{lastpage}
\jmlrheading{26}{2025}{1-\pageref{LastPage}}{3/25; Revised N.A.}{N.A.}{21-0000}{Jad Mounayer, Sebastian Rodriguez, Chady Ghnatios, Charbel Farhat, and Francisco Chinesta}

\ShortHeadings{Rank Reduction Autoencoders}{Mounayer et al.}
\firstpageno{1}

\begin{document}

\title{Rank Reduction Autoencoders}

\author{\name Jad Mounayer \email jad.mounayer@ensam.eu \\
       \addr PIMM Laboratory\\
       Arts et Metiers - ENSAM\\
       Paris, Bd. de l'Hopital, France
       \AND
       \name Sebastian Rodriguez \email sebastian.rodriguez\_iturra@ensam.eu\\
       \addr PIMM Laboratory\\
       Arts et Metiers - ENSAM\\
       Paris, Bd. de l'Hopital, France
        \AND
        \name Chady Ghnatios \email chady.ghnatios@unf.edu\\
       \addr University of North Florida\\
        1 UNF Dr., Jacksonville, USA
       \AND
        \name Charbel Farhat \email cfarhat@stanford.eu\\
       \addr Department of Aeronautics and Astronautics\\
       Stanford University\\
        450 Jane Stanford Way, Stanford, USA
        \AND
        \name Francisco Chinesta \email francisco.chinesta@ensam.eu\\
       \addr PIMM Laboratory\\
       Arts et Metiers - ENSAM\\
        Paris, Bd. de l'Hopital, France}
\editor{TBD}

\maketitle
\vspace{-0.2cm}
\begin{abstract}
The choice of an appropriate bottleneck dimension and the application of effective regularization are both essential for Autoencoders to learn meaningful representations from unlabeled data. In this paper, we introduce a new class of deterministic autoencoders, Rank Reduction Autoencoders (RRAEs), which regularize their latent spaces by employing a truncated singular value decomposition (SVD) during training. In RRAEs, the bottleneck is defined by the rank of the latent matrix, thereby alleviating the dependence of the encoder/decoder architecture on the bottleneck size. This approach enabled us to propose an adaptive algorithm (aRRAEs) that efficiently determines the optimal bottleneck size during training. We empirically demonstrate that both RRAEs and aRRAEs are stable, scalable, and reliable, as they do not introduce any additional training hyperparameters. We evaluate our proposed architecture on a synthetic data set, as well as on MNIST, Fashion MNIST, and CelebA. Our results show that RRAEs offer several advantages over Vanilla AEs with both large and small latent spaces, and outperform other regularizing AE architectures.
\end{abstract}


\newpage
\section{Introduction}

Compressed representations are crucial in many aspects of our daily lives. For instance, when recalling a familiar face, we do not consciously remember every detail but instead focus on key features such as the person's smile, eyes, or overall shape, and then mentally reconstruct the full image. Remarkably, this is possible even without knowing the person’s name or other personal details. In essence, our brains are capable of learning compressed yet meaningful representations of data, without relying on explicit labels.

In the past decade, neural networks have gained significant popularity due to their ability to approximate complex functions. Initially, neural networks were employed to map an input to a corresponding output. However, the advent of unsupervised learning has broadened the applicability of neural networks, enabling them to learn compressed representations of unlabeled data (as reviewed by \citealp{unsup_rev}). This process not only mirrors the way our brain works but also benefits supervised tasks, where the compressed representation itself can replace the original input in a neural network (for instance, \citealp{SAE,IAE}). One of the most widely used models for unsupervised learning is the autoencoder (reviewed by \citealp{AE_rev}). Autoencoders learn to encode input data into a compact representation, or ``latent space'', and then decode it back to reconstruct the original input. This process involves two key components: the encoder, which compresses the input into a lower-dimensional representation, and the decoder, which reconstructs the input from this compressed format. By minimizing the reconstruction error, the model learns to capture the most salient features of the data.

Vanilla autoencoders, in their original form, have served as the foundation for numerous studies. For example, \cite{speech_rec} developed them for speech recognition, and \cite{anomaly} used them for medical anomaly detection. A comprehensive review of various applications is provided by \cite{bank2021autoencoders}. However, vanilla autoencoders have two major limitations:

\begin{enumerate} 

    \item The size of the latent space determines the bottleneck (or compression level). Since both the encoder and decoder are dependent on the latent space dimension, this choice must be made prior to training and significantly influences the model’s performance. Selecting the appropriate latent space dimension can be challenging, particularly for complex, real-world data.
    
    \item The latent space of vanilla autoencoders is not regularized. Extensive research has been conducted to explore potential methods for regularizing autoencoders, with examples including the work of \cite{regularized_1}, \cite{regularized_2}, and \cite{regularized_3}. As demonstrated empirically later in this paper, non-regularized vanilla autoencoders can exhibit ``holes'' in their latent spaces, particularly when the latent space dimension is overestimated, leading to poor reconstructions by the decoder.
\end{enumerate}

To address the first issue, preprocessing techniques such as the principal component analysis (PCA) (\citealp{abdi2010principal}), the kernel PCA (\citealp{scholkopf1997kernel}), and the locally linear embedding (LLE) (\citealp{locally}) are used to estimate the intrinsic dimensionality of the data. While these methods can be effective, their performance tends to degrade as the intrinsic data dimensionality (or the rank of the data) increases. For example, these techniques often struggle to identify the correct bottleneck size for problems that involve more complex and/or local behaviors.

Several modifications to vanilla autoencoders have been proposed to allow the network to learn the bottleneck size while also regularizing the latent space. For instance, sparse autoencoders (\citealp{ng2011sparse}) promote a sparse representation of the data, ensuring that only a small number of features are active at any given time. Implicit Rank Minimizing Autoencoders (IRMAEs) (\citealp{implicit}) enable larger latent spaces while automatically determining the bottleneck by adding linear layers on top of the encoder. Contractive Autoencoders (CAEs) (\citealp{CAE}) enforce a contractive latent space around the training data.

While these methods improve regularization, they introduce the challenge of tuning additional hyperparameters. For example, the sparsity factor in sparse autoencoders, the number of layers in IRMAEs, or the loss term $\beta$ in CAEs must be manually adjusted. As demonstrated in Subsection \ref{HP}, incorrect values for these hyperparameters can degrade performance, often requiring extensive experimentation to identify optimal settings. Furthermore, both Sparse AEs and CAEs introduce additional terms to the loss function, complicating the optimization process and potentially worsening reconstruction errors.

In this paper, we introduce a novel class of autoencoders, \textit{Rank Reduction Autoencoders (RRAEs)}, designed to address the challenges outlined above. RRAEs overcome the bottleneck size issue while regularizing the latent space, all without the need for additional hyperparameters. The bottleneck is enforced in a strong manner within the architecture, so only the reconstruction loss needs to be minimized. The key innovation of RRAEs lies in determining the bottleneck through the singular values of the latent space matrix, rather than its dimensionality. As a result, RRAEs offer the following advantages:

\begin{itemize} 
 \item Since the singular values of the latent space determine the bottleneck, the design of the encoder and decoder is independent of the latent space dimension. This eliminates the need to predefine the bottleneck size. 
 \item Singular Value Decomposition (SVD) coefficients are regularized due to the orthogonality of the right singular vectors in the latent space, leading to more stable training without requiring additional regularization terms in the loss function. 
\item Singular values are sorted in a decreasing manner, meaning that the later singular values are less important for the model. This helps in creating a more meaningful latent space and enables the development of algorithms capable of efficiently identifying the optimal bottleneck dimension. \end{itemize}

Our experimental results show that when the bottleneck size is known in advance, RRAEs are stable and better regularized than Vanilla Autoencoders. We demonstrate their effectiveness in tasks such as image generation and sample interpolation. Furthermore, we introduce an adaptive algorithm that allows RRAEs to automatically determine the optimal bottleneck size during training. Our results show that RRAEs with the adaptive algorithm (aRRAEs) outperform other autoencoder architectures, all while eliminating the need for additional training efforts. Moreover, the algorithm is efficient and capable of identifying optimal bottleneck sizes that would be challenging to determine through manual testing.

\section{Autoencoders and Bottlenecks}\label{sec:AEs}
In this section, we define the notation for autoencoders and use it to introduce other autoencoder architectures of interest in the paper.

Let $X \in \mathbb{R}^{D \times N}$, where $N$ represents the number of samples and $D$ is the data dimension, be a set of unlabeled data for which we aim to learn a compact representation using an autoencoder. For a pre-specified bottleneck $k$, we define an encoding map $e: \mathbb{R}^{D \times N} \rightarrow \mathbb{R}^{k \times N}$ and a decoding map $d: \mathbb{R}^{k \times N} \rightarrow \mathbb{R}^{D \times N}$. A diabolo autoencoder can be expressed as the following two operations:

\begin{equation}\label{AE eq} Y = e(X), \qquad\qquad \tilde{X} = d(Y), \end{equation}

where $Y$ represents the \textbf{latent space}, the point at which the data is most compressed, and $e$ and $d$ are trainable neural networks. In practice, we impose the autoencoder to reconstruct the original data; thus, the loss function to minimize typically takes the form:

\begin{equation}\label{Loss AE} \mathcal{L}(X, \tilde{X}) = ||X - \tilde{X}||_2, \qquad \text{where} \qquad |\cdot|_2 \text{ denotes the } L2\text{-norm.} \end{equation}

As mentioned in the introduction, both the encoder and decoder in the Diabolo AE depend on the choice of bottleneck $k$, which must be selected a priori. In the following, we introduce other AE architectures that address this limitation.

\underline{Long AEs:} In this paper, we use the term ``Long AE'' to refer to autoencoders with long latent spaces. While a small latent space is beneficial due to the bottleneck constraint, choosing any $L \neq k$ as the latent space dimension prevents the autoencoder from learning the identity function and may instead learn certain features of the data. Long AEs are defined using equations \eqref{AE eq} and \eqref{Loss AE}, but with $e: \mathbb{R}^{D \times N} \rightarrow \mathbb{R}^{L \times N}$, for any $L > k$. The advantage of this architecture is that it mitigates the dependency on a bottleneck by replacing it with the task of selecting any value greater than $k$, which is typically easier by choosing a sufficiently large $L$. However, the main limitation of Long AEs is that their latent spaces may have ``holes''. In other words, while exploring the latent space, we often do not remain on the data manifold, which complicates tasks such as interpolation and data generation, as latent space coefficients may not carry the same meaning as in a Diabolo AE with a carefully chosen bottleneck.

\underline{Sparse AEs:} Sparse autoencoders are an enhancement of Long AEs, where the latent space is constrained to be sparse, functioning as a bottleneck and thereby leading to more meaningful latent representations. As presented by \cite{ng2011sparse}, Sparsity is achieved by enforcing the latent space to be positive and encouraging latent values to remain as small as possible. Sparse AEs are formulated as:

\begin{equation}\nonumber Y = \sigma(e(X)), \qquad\qquad \tilde{X} = d(Y), \end{equation}

where $\sigma$ is the sigmoid activation function, ensuring that $Y$ remains within the range $[0, 1]$. Additionally, the loss function incorporates extra terms:

\begin{equation}\label{sparse_loss} \mathcal{L}(X, \tilde{X}) = ||X - \tilde{X}||_2 \quad + \beta \rho \log\left(\frac{\rho}{\text{mean}(Y)} + \epsilon\right) + \beta(1 - \rho) \log\left(\frac{1 - \rho}{1 - \text{mean}(Y) + \epsilon}\right), \end{equation}

where $\epsilon$ is a small constant to prevent invalid values, $\rho$ is the sparsity parameter that controls the desired sparsity of the latent space, and $\beta$ is a hyperparameter that scales the sparsity term relative to the reconstruction error. At convergence, the additional terms in the loss function ensure that the mean of the latent space is close to the sparsity $\rho$. While enforcing sparsity improves the consistency of long latent spaces, our results indicate that both $\rho$ and $\beta$ should be carefully selected to achieve optimal performance. This typically involves a trial-and-error process, as the best values of $\rho$ and $\beta$ depend on factors such as the data, model architecture, and training process. Furthermore, while sparsity helps refine the latent space, it does not provide a compressed representation for each data sample, as is the case with the Diabolo autoencoder (in other words, we do not obtain a compressed representation in $\mathbb{R}^{k}$ for every sample in $\mathbb{R}^{D}$). Additionally, equation \eqref{sparse_loss} includes multiple losses, which can lead to a more complex optimization process, leading to worse reconstruction errors.

\underline{Implicit Rank-Minimizing AEs (IRMAEs):} \cite{chow:68} introduced a simple and effective method to enhance the behavior of Long AEs. The core idea of IRMAEs is to enforce the latent space to have a low matrix rank, which can mimic the behavior of a bottleneck if the rank is sufficiently small. The authors proposed adding linear layers on top of the encoder, demonstrating that this encourages the latent space to exhibit low matrix rank characteristics, resulting in a more compact representation. Thus, IRMAEs are defined as a Long AE in equations \eqref{AE eq} and \eqref{Loss AE}, but with $l$ linear layers added to the encoding map $e$. Although this enhancement improves the use of AEs with long latent spaces, the value of $l$ must be carefully chosen to achieve good results. Since $l$ depends on factors such as the data, model architecture, and training process, training IRMAEs may require several iterations to find a suitable latent space with low rank that approximates the intrinsic bottleneck of the data. Additionally, the low rank of the latent space is not strictly enforced, leading to many singular values being close to zero but not exactly zero. Consequently, it is not guaranteed that we will obtain a bottleneck representation of dimension $k < D$.

\underline{Contractive AEs (CAEs):} \cite{CAE} proposed a different method to enhance the robustness of the latent features. Since the dependence of the latent space of an autoencoder on its inputs can be quantified by the magnitude of the Jacobian of the nonlinear mapping, CAEs penalize large values of the Jacobian by adding a regularization term:

\begin{equation}\nonumber \mathcal{L}(X, \tilde{X}) = ||X - \tilde{X}||2 \quad + \beta \sum_{ij}\left(\frac{\partial h_j(X)}{\partial X_i}\right)^2, \end{equation}

where $h$ represents the latent space, and subscripts represent the element index. The term on the right-hand side represents the Frobenius norm of the Jacobian. Penalizing this value encourages the latent space to be contractive around the training data, leading to more robust feature representations. Contractive AEs face similar challenges to Sparse AEs. First, if the bottleneck is not specified a priori, a compressed representation of the data cannot be obtained. Second, CAEs include a hyperparameter $\beta$, which must be tuned based on the loss values. Finally, the optimization process is more complex, as the reconstruction error is no longer the sole objective.

In summary, autoencoders perform well when the bottleneck is correctly chosen. However, assuming prior knowledge of this parameter can be overly restrictive. Several AE architectures have been proposed to address this issue. Long AEs suffer from holes in their latent spaces, while Sparse, Implicit Rank Minimizing, and Contractive Autoencoders introduce new hyperparameters to tune in place of the bottleneck $k$. Moreover, none of these architectures yield a compressed representation of the data. In other words, after training, we cannot represent the data $X \in \mathbb{R}^{D \times N}$ using a bottleneck of reduced dimension $\mathbb{R}^{k \times N}$ with $k < D$.

In the following, we introduce a novel class of autoencoders, Rank Reduction Autoencoders (RRAEs), which alleviate the dependency of the model architecture on the bottleneck $k$, without introducing an additional hyperparameter or a secondary term in the loss function. Furthermore, RRAEs, like Diabolo AEs, allow us to obtain a compressed representation of the data once training is complete.

\section{The Singular Value Decomposition (SVD)}
The concept of Rank Reduction Autoencoders depends on the truncated SVD. Accordingly, we define in this section the required SVD notations and concepts to understand the idea behind RRAEs and their convergence. \cite{yuji} offers an introduction to the SVD along with important proofs. For any matrix $Y \in \mathbb{R}^{L \times N}$, we can define the singular value decomposition as,

\begin{equation}\label{SVD orig}
  Y = U\Sigma V^T, \qquad \text{with: }\qquad U \in \mathbb{R}^{L\times k}, \qquad \Sigma \in \mathbb{R}^{k\times k}, \qquad \text{and,} \qquad V^T \in \mathbb{R}^{k\times N},
\end{equation}
with $k$ being the rank of the matrix, $U^TU = V^TV = I_k$, and $\Sigma$ a diagonal matrix containing the nonnegative singular values sorted in decreasing order. Writing $U = [u_1, \ldots, u_k]$, $V = [v_1, \ldots, v_k]$, and $\Sigma=\text{diag}(\sigma_1, \ldots, \sigma_k)$, we can define the SVD as the sum of rank-1 updates as follows,

\begin{equation}\label{svd sum}
  \displaystyle Y = \sum_{i=1}^{k}u_i\sigma_iv_i^T.
\end{equation}
Moreover, writing $Y = [y_1, \ldots, y_N]$, and $V^T=[r_1, \ldots, r_N]$, we can define each column of $Y$ using the SVD as,
\begin{equation}\label{SVD with alpha}
  y_i = U \Sigma r_i = U \alpha_i, \qquad \text{with,} \qquad \alpha_i = \Sigma r_i, \qquad \forall i \in [1, N].
\end{equation}
Equivalently, defining $\alpha = [\alpha_1, \ldots, \alpha_N]$, we can write equation \eqref{svd sum} in matrix form as $Y = U\alpha$. Note that no matter which column $Y$ is to be represented, the entire matrix $U$ is used in equation \eqref{SVD with alpha}. Accordingly, if $U$ itself is known a priori, each column of $Y$ can be represented by its corresponding $\alpha_i \in \mathbb{R}^k$. Throughout this paper, we will call the columns of matrix $U$ the basis vectors, and $U$ itself the basis. On the other hand, we call the $\alpha$ values the coefficients representing each column of matrix $Y$ in the basis $U$.

The discussion above shows how the SVD is a way of writing any matrix $Y$ as a set of basis vectors multiplied by specific coefficients per column. However, when the rank $k$ of a matrix is large, the sum in equation \eqref{svd sum} will contain multiple terms, and the length of vectors $\alpha_i$ becomes larger. Since the singular values $\sigma_i$ are sorted in decreasing order, the SVD also allows us to find optimal approximations of $Y$ using only the $k^*<k$ most important singular values. This is known as the truncated SVD, which allows us to write the rank-$k^*$ approximation of $Y$ as follows,
\begin{equation}\label{trunc SVD}
  Y \approx Y^{(k^*)} = \sum_{i=1}^{k^*}u_i\sigma_iv_i = \underbrace{\begin{bmatrix}
         &        &        \\
         &        &        \\
     u_1 & \ldots & u_{k^*}\\
     &        &     \\
     &        &     \\
  \end{bmatrix}}_{:= \, \, U^{(k^*)}}
  \underbrace{\begin{bmatrix}
  \sigma_1 &        &        \\
& \ddots & \\
&        &    \sigma_{k^*}    \\
\end{bmatrix}}_{:= \, \, \Sigma^{(k^*)}}
\underbrace{\begin{bmatrix}
  &    &    &   v_1^T    &    & & \\
 & & & \vdots &   & &\\
&    &    &  v_{k^*}^T   &   & & \\
\end{bmatrix}}_{:= \, \, V^{T(k^*)}}.
\end{equation}
As shown in equation \eqref{trunc SVD}, the truncated SVD simply consists of taking fewer rank-1 updates in equation \eqref{svd sum}, or fewer columns, diagonal values, and rows from $U$, $\Sigma$, and $V^T$, respectively in equation \eqref{SVD orig}.

Note that because of the orthonormality of the basis $U$, the truncated sum in equation \eqref{trunc SVD} can also be written as a projection on the truncated basis $U^{(k^*)}$ as follows,

\begin{equation}\label{trunc projection}
  Y \approx Y^{(k^*)} = U^{(k^*)}\underbrace{\left(U^{(k^*)}\right)^T Y}_{\alpha^{(k^*)}}
\end{equation}
Here, $\alpha_i^{(k^*)} \in \mathbb{R}^{k^*}$ is a vector containing the coefficients that represent each column $Y_i$. Note the following,

\begin{observation}\label{bottleneck}
  Let $Y \in \mathbb{R}^{L \times N}$ be a matrix with rank $k$, and let $Y^{(k^*)}$ be the matrix resulting from the truncated SVD of $Y$ with $k^* \leq k$, based on equations \eqref{SVD with alpha} and \eqref{trunc SVD}, if the basis $U^{(k^*)}$ is found a priori, each column of  $Y_i^{(k^*)}$ can be represented by its corresponding coefficients $\alpha_i^{(k^*)} \in \mathbb{R}^{k^*}$.
\end{observation}
In other words, truncating an SVD at $k^*$ is equivalent to finding a bottleneck of size $k^*$ if the basis is known.

\section{Rank Reduction Autoencoders (RRAEs) - Fixed Bottleneck}\label{sec:RRAE}
The truncated singular value decomposition (SVD) is a widely used technique for data dimensionality reduction. It serves as the foundation for several well-established methods, such as the principal component analysis (PCA) and the proper orthogonal decomposition (POD). However, these techniques are inherently linear, which limits their effectiveness in compressing data when the underlying structure is nonlinear. Building upon the advantageous properties of the SVD, we introduce Rank Reduction Autoencoders (RRAEs), which combine the linear dimensionality reduction of truncated SVD with the nonlinear nature of an Autoencoder's latent space. For simplicity, we first present the model in its basic form, where the bottleneck size is predefined and training is performed without batch processing.

\subsection{RRAEs Without Batches}\label{subsec: RRAE_fixed_whole}

This methodology is based on the truncated SVD formulations in equations \eqref{trunc SVD} and \eqref{trunc projection}, as illustrated in the top portion of Figure \ref{fig:RRAE}. The key idea behind RRAEs is to impose a bottleneck on the singular values within the latent space matrix, rather than restricting the dimensionality of the matrix itself.

\tikzset{every picture/.style={line width=0.75pt}} 
\begin{figure}[!t]
\begin{tikzpicture}[x=0.75pt,y=0.75pt,yscale=-1,xscale=1]

\draw    (111.67,84.67) -- (167.67,66.42) ;
\draw    (111.67,114) -- (167.67,131) ;
\draw    (167.67,66.42) -- (167.67,131.33) ;
\draw    (111.67,84.67) -- (111.67,114) ;
\draw    (391,66.42) -- (451, 84.67) ;
\draw    (450, 114) -- (391, 131) ;
\draw    (391, 66.42) -- (391, 131) ;
\draw    (451, 84.67) -- (451, 114) ;
\draw    (281,102.22) -- (281,160.67) ; 
\draw   [->] (271,102) -- (291,102) ;
\draw   [<->] (25.8,78.22) -- (25.8,120.67) ;
\draw  [<->]  (185.6,62.11) -- (185.6, 135) ;
\draw  [color={rgb, 255:red, 208; green, 2; blue, 27 }  ,draw opacity=1 ] (248.67,163.86) .. controls (248.67,158.81) and (252.76,154.72) .. (257.81,154.72) .. controls (262.85,154.72) and (266.94,158.81) .. (266.94,163.86) .. controls (266.94,168.91) and (262.85,173) .. (257.81,173) .. controls (252.76,173) and (248.67,168.91) .. (248.67,163.86) -- cycle ;
\draw [color={rgb, 255:red, 208; green, 2; blue, 27 }  ,draw opacity=1, -> ]   (248.67,163.86) -- (210.37,177) ;
\draw  [color={rgb, 255:red, 208; green, 2; blue, 27 }  ,draw opacity=1 ] (179,277.78) .. controls (179,270.17) and (185.17,264) .. (192.78,264) .. controls (200.39,264) and (206.56,270.17) .. (206.56,277.78) .. controls (206.56,285.39) and (200.39,291.56) .. (192.78,291.56) .. controls (185.17,291.56) and (179,285.39) .. (179,277.78) -- cycle ;
\draw [<-] [color={rgb, 255:red, 208; green, 2; blue, 27 }  ,draw opacity=1 ]   (193.78, 309) -- (193.78, 291.26) ;

\draw    (154.13,419.67) -- (210.13,401.42) ;
\draw    (154.13,449) -- (210.13,466) ;
\draw    (210.13,401.42) -- (210.13,466.33) ;
\draw    (154.13,419.67) -- (154.13,449) ;

\draw    (355.47,399.92) -- (414.97,419.42) ;
\draw    (414.97,448.75) -- (354.97,468.92) ;
\draw    (355.47,399.92) -- (354.97,468.92) ;
\draw    (414.97,419.42) -- (414.97,448.75) ;

\draw  [->]  (265.8, 432.67) -- (285, 432.67) ;

\draw [color={rgb, 255:red, 208; green, 2; blue, 27 }, draw opacity=1] (329, 433) circle (22);

\draw [color={rgb, 255:red, 208; green, 2; blue, 27 }  ,draw opacity=1 , ->]   (327.11, 455.11) -- (294.57, 500.4) ;

\draw (116.33,94.67) node [anchor=north west][inner sep=0.75pt]  [font=\footnotesize] [align=left] {Encoder};
\draw (398.67,94.67) node [anchor=north west][inner sep=0.75pt]  [font=\footnotesize] [align=left] {Decoder};
\draw (5.33,95.27) node [anchor=north west][inner sep=0.75pt]    {$D$};
\draw (170.8,91.8) node [anchor=north west][inner sep=0.75pt]    {$L$};
\draw (248.2,158.33) node [anchor=north west][inner sep=0.75pt]    {$\displaystyle \sum _{l=1}^{k^{*}} \sigma _{l} u_{l} v_{l}^{T}$};
\draw (395.4,176.07) node [anchor=north west][inner sep=0.75pt]    {$\mathcal{L} =\ \| \ X-\tilde{X} \ \Vert _{2}$};
\draw (25.33,11.67) node [anchor=north west][inner sep=0.75pt]   [align=left] {\underline{{\large Training:}}};
\draw (29.67,349.67) node [anchor=north west][inner sep=0.75pt]   [align=left] {{\large \underline{Inference (one new sample):}}};
\draw (75.83, 186.83) node [anchor=north west][inner sep=0.75pt]  [font=\large,color={rgb, 255:red, 208; green, 2; blue, 27 }  ,opacity=1 ] [align=left] {{\large Enforcing bottleneck}};
\draw (25,226) node [anchor=north west][inner sep=0.75pt]   [align=left] {\underline{{\large Post-training:}}};
\draw (185.3,268.13) node [anchor=north west][inner sep=0.75pt]    {$U_{f} \Sigma_{f} V_{f}^{T} =\ SVD_{k^*}(\text{Encoder}(X_{train}))$};
\draw (128.5,308.5) node [anchor=north west][inner sep=0.75pt]  [font=\scriptsize,color={rgb, 255:red, 208; green, 2; blue, 27 }  ,opacity=1 ] [align=left] {{\large Find Training basis $\displaystyle \in \mathbb{R}^{^{L\times k}}$}};
\draw (25.8,58.62) node [anchor=north west][inner sep=0.75pt]  [font=\footnotesize]  {$\begin{bmatrix}
 &  & \\
X_{1} \!\!\!  & \dotsc  \!\!\! & X_{N}\\
 &  & 
\end{bmatrix} \quad \quad \quad \ \quad \quad \ \ \begin{bmatrix}
 &  & \\
 &  & \\
Y_{1}\! & \dotsc  & \! Y_{N}\\
 &  & \\
 &  & 
\end{bmatrix} \ \ \ \ \begin{bmatrix}
 &  & \\
 &  & \\
Y_{1}^{^{\left( k^{*}\right)}} \!\!\!\!\!\!\! & \dotsc  \!\!\!\!\! & Y_{N}^{^{\left( k^{*}\right)}}\\
 &  & \\
 &  & 
\end{bmatrix} \quad \quad \quad \ \quad \ \ \begin{bmatrix}
 &  & \\
\tilde{X}_{1} \!\!\! & \dotsc  & \!\!\! \tilde{X}_{N}\\
 &  & 
\end{bmatrix}$};
\draw (159, 428.67) node [anchor=north west][inner sep=0.75pt]  [font=\footnotesize] [align=left] {Encoder};
\draw (361, 428.67) node [anchor=north west][inner sep=0.75pt]  [font=\footnotesize] [align=left] {Decoder};
\draw (104.97,380.12) node [anchor=north west][inner sep=0.75pt]  [font=\footnotesize]  {$\begin{bmatrix}
 \\
 \\
X_{new}\\
\\
 \\
\end{bmatrix} \quad \quad \quad \quad \quad \begin{bmatrix}
 \\\\
 \\
\ Y_{new} \ \\
 \\\\
 \\
\end{bmatrix} \quad \quad \ U_{f} U_{f}^{T} Y_{new} \quad \quad \quad \quad \quad  \begin{bmatrix}
 \\
 \\
\tilde{X}_{new} \\
 \\
 \\
\end{bmatrix}$};
\draw (213.5, 502.83) node [anchor=north west][inner sep=0.75pt]  [font=\scriptsize,color={rgb, 255:red, 208; green, 2; blue, 27 }  ,opacity=1 ] [align=left] {{\large Bottleneck coefficients $\displaystyle \in \mathbb{R}^{k \times 1}$}};
\end{tikzpicture}
\caption{Schematic illustrating RRAE's architecture when trained without batches.}
\label{fig:RRAE}
\end{figure}

Let $X \in \mathbb{R}^{D\times N}$ be a set of $N$ samples of unlabeled data that we would like to represent with an Autoencoder, and a bottleneck of size $k^*$. In this case, we assume that the input is two-dimensional without loss of generality, since inputs of higher dimensions would simply require flattening/reshape layers inside the encoder/decoder respectively to get an equivalent behavior. Similarly to the Autoencoders previously presented in Section \ref{sec:AEs}, RRAEs consist of both an encoding and a decoding map, $e: \mathbb{R}^{D\times N} \rightarrow \mathbb{R}^{L\times N}$, and $d: \mathbb{R}^{L\times N} \rightarrow \mathbb{R}^{D\times N}$ where $L$ is the dimension of the latent space which has to be bigger than the bottleneck (so $L > k^*$). Note that the only restriction on $L$ is $L> k^*$, it can be either bigger or smaller than the original data dimension $D$. During training, an RRAE encodes the data into its latent matrix $Y \in \mathbb{R}^{L\times N}$, performs its truncated SVD of rank $k^*$ (the specified bottleneck), and only passes the truncated matrix to the decoder.
The advantages of using a truncated SVD in the latent space are as follows:
\begin{itemize}
  \item The bottleneck enforced by the truncated SVD is of dimension $k^*$. Note that in this setup, each column of matrix $Y$ represents a sample of the data. As discussed in Observation \ref{bottleneck}, each column of the truncated latent matrix $Y^{(k^*)}$ can be represented by a bottleneck of size $k^*$. Accordingly, even though the input of the decoder is of shape $L \times N$, the decoder is receiving a compressed version of the data of size $k^*$, multiplied by a larger basis that's common between all data samples.
  \item The truncated SVD allows us to impose the bottleneck discussed in the previous point without any additional loss terms or hyperparameters. As shown in the top part of Figure \ref{fig:RRAE}, the loss simply consists of the reconstruction error.
  \item The bottleneck coefficients are regularized. Since the coefficients constitute the truncated right singular vector, the coefficients have to be both normalized and orthonormal to each other. This property allows RRAEs to be stable during training without requiring any regularization on the network's weights and biases.
  \item The bottleneck coefficients are arranged by their importance. These coefficients are made up of singular values multiplied by normalized vectors, with the singular value indicating the coefficient's significance. Since singular values are positive and ordered from largest to smallest, we can assume that the first few coefficients are the most important. This property is helpful when designing adaptive algorithms (see Section \ref{sec: aRRAE} for more details).
  \item The dependence of each sample's latent space on other samples. While in other Autoencoder architectures, each sample is independent of the others, the truncated SVD takes all $D$ samples (since no batching is done yet), and hence each column of the truncated latent space $Y^{(k^*)}$ depends on the other samples as well. This helps RRAEs in learning a more global behavior instead of treating each sample independently.
\end{itemize}
While the truncated SVD brings many advantages during training, inference is not straightforward. On the one hand, performing the truncated SVD during inference limits us to passing multiple samples at the same time and makes evaluation sample-dependent. For instance, if only one sample is given to the network, the truncated SVD of a column vector is simply the column vector itself (hence no compressed representation). On the other hand, if a new (or test) set is passed to the network, the truncated SVD of its latent matrix will find a different basis than the one used in training,which is counter-intuitive. As stated in Observation \ref{bottleneck}, we can only define a compressed representation if the basis is known a priori. Accordingly, after finding a basis $U_f$ for the training set, the truncated SVD in the latent space during inference is replaced by a projection onto the basis $U_f$, as illustrated in the expression of the SVD in equation \eqref{trunc projection}. The full procedure when training on a full data set with a fixed bottleneck is as follows:
\begin{enumerate}
  \item Train the encoder/decoder by passing the entire data set into the network, and optimizing the reconstruction loss with a truncated SVD of rank $k^*$ in the latent space.
  \item Once the model is trained, the entire training data set is passed one last time into the encoder to find its latent space, without any training backpropagation. A final truncated SVD is performed to find and save the final truncated basis $U_f := U^{(k^*)}$.
  \item For inference, instead of performing a truncated SVD, we project the latent space onto the basis $U_f$ as shown in the bottom of Figure \ref{fig:RRAE}. Basically, during training, the truncated SVD defined in equation \eqref{trunc SVD} is used as it offers multiple advantages, as previously described. Inference, on the other hand, is based on equation \eqref{trunc projection} where we use the basis found during training to evaluate the model over new samples.
\end{enumerate}

While the steps listed above work well when the entire data set is used in training (so without batches), batching is problematic since we would have a different basis $U_f$ for each batch. This is explored in the next subsection.

\subsection{RRAEs With Batches} \label{subsec: RRAE_fixed_batches}
Batching plays a crucial role in machine learning. When working with large data sets, training without batches is not feasible. In this subsection, we explain why RRAEs converge when using batches and how to compute the final training basis $U_f$ when batching is applied. We continue with the same unlabeled data set $X \in \mathbb{R}^{D\times N}$ , which is now divided into batches before being input to the network. For a given batch size $B$, the input to the encoder has dimensions ($D \times B$) and the corresponding latent matrix is of size ($L \times B$). Since the truncated SVD depends on all the columns in a matrix, each batch will have its own basis. But are these bases related? Specifically, can we find a common basis $U_f$ that allows us to reduce the entire data set to a bottleneck once training is finished?

In the following, we present reasons to support the following observation,
\begin{observation}\label{obs: batch}
  Let $L$ be a chosen latent space dimension, and $X\in \mathbb{R}^{D\times N}$ be a set of unlabeled data with at least $n \geq L$ linearly independent samples. If the RRAE of latent size $L$ is trained until convergence using batches, the bases found for each batch are close to being linear combinations of each others.
\end{observation}
In the above, \emph{convergence} is defined by the following two events,
\begin{enumerate}
  \item Whenever a sample $X_i$ is passed as an input to the network, its reconstruction $\tilde{X}_i$ is close to $X_i$.
  \item A good encoder/decoder couple has been found using backpropagation. Accordingly, the encoder/decoder do not vary much anymore from an epoch to another.
\end{enumerate}

To back up Observation \ref{obs: batch}, let $X_i$ be a column (or a sample) of $X$ and assume we have an RRAE trained until convergence. Recall that the truncated SVD in the latent space depends on all samples selected in the batch. Accordingly, if after convergence, $X_i$ is passed at two different epochs along with a different combination of samples due to random batching, the truncated representation $Y_i^{(k^*)}$ can be expressed differently (check Figure \ref{fig:batches} for an example of $i=2$ and a batch size of 3). Specifically, for two different batches containing the same sample $X_i$, we can write,

\begin{figure}[!t]
  \begin{tikzpicture}[x=0.75pt,y=0.75pt,yscale=-1,xscale=1]
  
  \draw    (111.67,84.67) -- (167.67,66.42) ;
  \draw    (111.67,114) -- (167.67,131) ;
  \draw    (167.67,66.42) -- (167.67,131.33) ;
  \draw    (111.67,84.67) -- (111.67,114) ;

  \draw    (400,66.42) -- (460, 84.67) ;
  \draw    (460, 114) -- (400, 131) ;
  \draw    (400, 66.42) -- (400, 131) ;
  \draw    (460, 84.67) -- (460, 114) ;
  \draw    (283,102.22) -- (283,160.67) ; 

  \draw   [->] (273,102) -- (293,102) ;
  \draw   [<->] (25.8,78.22) -- (25.8,120.67) ;
  \draw  [<->]  (185.6,62.11) -- (185.6, 135) ;

  \draw (116.33,94.67) node [anchor=north west][inner sep=0.75pt]  [font=\footnotesize] [align=left] {Encoder};
  \draw (404.67,94.67) node [anchor=north west][inner sep=0.75pt]  [font=\footnotesize] [align=left] {Decoder};
  \draw (5.33,95.27) node [anchor=north west][inner sep=0.75pt]    {$D$};
  \draw (170.8,91.8) node [anchor=north west][inner sep=0.75pt]    {$L$};
  \draw (248.2,158.33) node [anchor=north west][inner sep=0.75pt]    {$\displaystyle \sum _{l=1}^{k^{*}} \sigma _{l} u_{l} v_{l}^{T}$};
  \draw (25.33,11.67) node [anchor=north west][inner sep=0.75pt]   [align=left] {\underline{{\large $X_2$ at epoch $i$ after convergence:}}};
  \draw (29.8,58.62) node [anchor=north west][inner sep=0.75pt]  [font=\footnotesize]  {$\arraycolsep=1.4pt\left[\begin{array}{c|c|c}
  \cellcolor{olive!20} & \cellcolor{yellow!20} & \cellcolor{green!20}\\
  \cellcolor{olive!20} X_{1}  &\cellcolor{yellow!20} X_{2} & \cellcolor{green!20} X_{3}\\
  \cellcolor{olive!20} & \cellcolor{yellow!20}  & \cellcolor{green!20} 
  \end{array}\right] \quad \quad \quad \ \quad \quad \quad \, \arraycolsep=4pt\left[\begin{array}{c|c|c}
    \cellcolor{olive!20} & \cellcolor{yellow!20} & \cellcolor{green!20}\\
    \cellcolor{olive!20} & \cellcolor{yellow!20} & \cellcolor{green!20}\\
    \cellcolor{olive!20}  Y_{1}\! & \cellcolor{yellow!20} Y_{2}  & \! \cellcolor{green!20} Y_{3}\\
    \cellcolor{olive!20} & \cellcolor{yellow!20} & \cellcolor{green!20}\\
    \cellcolor{olive!20} & \cellcolor{yellow!20} & \cellcolor{green!20}
  \end{array}\right] \ \ \ \ \arraycolsep=0.1pt\left[\begin{array}{c|c|c}
    \cellcolor{green!20}  \! & \cellcolor{olive!20}  & \! \cellcolor{yellow!20}\\[-0.5pt]
    \cellcolor{olive!20} & \cellcolor{yellow!20} & \cellcolor{green!20}\\[-0.5pt]
    \cellcolor{yellow!20} Y_{1}^{(k^*)} & \cellcolor{green!20} Y_{2}^{(k^*)}  & \cellcolor{olive!20} Y_{3}^{(k^*)} \\[-0.5pt]
    \cellcolor{green!20} & \cellcolor{olive!20} & \cellcolor{yellow!20}\\[-0.5pt]
    \cellcolor{olive!20} & \cellcolor{yellow!20} & \cellcolor{green!20}\\[-0.5pt]
  \end{array}\right]  \quad \quad \quad \ \quad \quad \arraycolsep=1.4pt\left[\begin{array}{c|c|c}
    &  & \\
    \tilde{X}_{1}  & \tilde{X}_{2} & \tilde{X}_{3}\\
    &   & 
    \end{array}\right]$};

    \draw    (111.67,304.67) -- (167.67,286.42) ;
    \draw    (111.67,334) -- (167.67,351) ;
    \draw    (167.67,286.42) -- (167.67,351.33) ;
    \draw    (111.67,304.67) -- (111.67,334) ;

    \draw    (400,286.42) -- (460,304.67) ;
    \draw    (460,334) -- (400,351) ;
    \draw    (400,286.42) -- (400,351) ;
    \draw    (460,304.67) -- (460,334) ;
    \draw    (283,322.22) -- (283,380.67) ; 

    \draw   [->] (273,322) -- (293,322) ;
    \draw   [<->] (25.8,298.22) -- (25.8,340.67) ;
    \draw  [<->]  (185.6,282.11) -- (185.6,355) ;

    \draw (116.33,314.67) node [anchor=north west][inner sep=0.75pt]  [font=\footnotesize] [align=left] {Encoder};
    \draw (404.67,314.67) node [anchor=north west][inner sep=0.75pt]  [font=\footnotesize] [align=left] {Decoder};
    \draw (5.33,315.27) node [anchor=north west][inner sep=0.75pt]    {$D$};
    \draw (170.8,311.8) node [anchor=north west][inner sep=0.75pt]    {$L$};
    \draw (248.2,378.33) node [anchor=north west][inner sep=0.75pt]    {$\displaystyle \sum _{l=1}^{k^{*}} \sigma _{l} u_{l} v_{l}^{T}$};
    \draw (25.33,231.67) node [anchor=north west][inner sep=0.75pt]   [align=left] {\underline{{\large $X_2$ at epoch $j$ after convergence:}}};
    \draw (29.8,278.62) node [anchor=north west][inner sep=0.75pt]  [font=\footnotesize]  {$\arraycolsep=1.4pt\left[\begin{array}{c|c|c}
    \cellcolor{cyan!20} & \cellcolor{yellow!20} & \cellcolor{orange!20}\\
      \cellcolor{cyan!20} X_{8}  &\cellcolor{yellow!20} X_{2} & \cellcolor{orange!20} X_{9}\\
      \cellcolor{cyan!20} & \cellcolor{yellow!20}  & \cellcolor{orange!20} 
      \end{array}\right] \quad \quad \quad \ \quad \quad \quad \, \arraycolsep=4pt\left[\begin{array}{c|c|c}
      \cellcolor{cyan!20} & \cellcolor{yellow!20} & \cellcolor{orange!20}\\
      \cellcolor{cyan!20} & \cellcolor{yellow!20} & \cellcolor{orange!20}\\
      \cellcolor{cyan!20}  Y_{8}\! & \cellcolor{yellow!20} Y_{2}  & \! \cellcolor{orange!20} Y_{9}\\
      \cellcolor{cyan!20} & \cellcolor{yellow!20} & \cellcolor{orange!20}\\
      \cellcolor{cyan!20} & \cellcolor{yellow!20} & \cellcolor{orange!20}
      \end{array}\right] \ \ \ \ \arraycolsep=0.1pt\left[\begin{array}{c|c|c}
      \cellcolor{orange!20}  \! & \cellcolor{cyan!20}  & \! \cellcolor{yellow!20}\\[-0.5pt]
      \cellcolor{cyan!20} & \cellcolor{yellow!20} & \cellcolor{orange!20}\\[-0.5pt]
      \cellcolor{yellow!20} Y_{8}^{(k^*)} & \cellcolor{orange!20} Y_{2}^{(k^*)}  & \cellcolor{cyan!20} Y_{9}^{(k^*)} \\[-0.5pt]
      \cellcolor{orange!20} & \cellcolor{cyan!20} & \cellcolor{yellow!20}\\[-0.5pt]
      \cellcolor{cyan!20} & \cellcolor{yellow!20} & \cellcolor{orange!20}\\[-0.5pt]
      \end{array}\right]  \quad \quad \quad \ \quad \quad \arraycolsep=1.4pt\left[\begin{array}{c|c|c}
      &  & \\
      \tilde{X}_{8}  & \tilde{X}_{2} & \tilde{X}_{9}\\
      &   & 
      \end{array}\right]$};
  \draw [<->] (345, 132) -- (345, 280) ;
  \draw (345, 192) node [anchor=north west] [font=\LARGE] {$\neq$};

  \end{tikzpicture}
  \caption{Schematic illustrating why a latent vector $Y_i$ for a sample $X_i$ can be written differently when $X_i$ is processed with different batches. The colors are only to illustrate the dependence on a column.}
  \label{fig:batches}
  \end{figure}
  
\begin{equation}\label{eq: mult batch}
  \begin{cases}
    \left(Y_i^{(k^*)}\right)^{batch_1} = U_f^{batch_1} \left(U_f^{batch_1}\right)^T Y_i, \\[0.4cm]
    \left(Y_i^{(k^*)}\right)^{batch_2} = U_f^{batch_2} \left(U_f^{batch_2}\right)^T Y_i.
  \end{cases}
\end{equation}
Where $Y_i$ is the same between both batches since we assumed that the encoder had already converged. We can also write,
\begin{equation}\label{eq: with dec}
\tilde{X}_i^{batch_1}\approx\tilde{X}_i^{batch_2} \implies \text{Decoder}\left(\left(Y_i^{(k^*)}\right)^{batch_1}\right) \approx \text{Decoder}\left(\left(Y_i^{(k^*)}\right)^{batch_2}\right) . 
\end{equation}
Combining equations \eqref{eq: mult batch} and \eqref{eq: with dec}, and since we assume that the decoder doesn't vary much after convergence, we can write,
\begin{equation}\label{eq:uut}
  U_f^{batch_1} \left(U_f^{batch_1}\right)^T Y_i \approx U_f^{batch_2} \left(U_f^{batch_2}\right)^T Y_i, \quad \text{or,}\quad AY_i\approx[0]_L,
\end{equation}
with $A = U_f^{batch_1} \left(U_f^{batch_1}\right)^T - U_f^{batch_2} \left(U_f^{batch_2}\right)^T$. Note the following Lemma,
\begin{lemma}
  Let $A \in \mathbb{R}^{L \times L}$, if there exists at least $L$ linearly independent samples $Y_i\in\mathbb{R}^{L}$ for which $AY_i = 0$, then $A=[0]_{L\times L}$.
\end{lemma}

\begin{proof} 
By the rank-nullity theorem (see \citealp{lay2012linear}), $L = \text{Rank}(A) + \text{Nullity}(A)$. If there exists at least $L$ linearly independent samples such as $AY_i=0$, $\text{Nullity}(L) = L$, and hence the rank of $A$ is equal to zero. Note that if a matrix is of rank zero, all of its singular values are zero so using the full SVD of $A$, we can write, $A$ = $U[0]_{L\times L} V^T = [0]_{L\times L}.$
\end{proof}

Since we assume that there exists $n \geq L$ linearly independent samples, based on equation \eqref{eq:uut}, we can write,
\begin{equation}\nonumber
  A \approx [0]_{L\times L} \quad \implies \quad U_f^{batch_1}\left(U_f^{batch_1}\right)^T \approx U_f^{batch_2} \left(U_f^{batch_2}\right)^T,
\end{equation}
In other words, we can write, $U_f^{batch_1} \approx U_f^{batch_2} W$, or more generally for any two batches $j$ and $k$,
\begin{equation}\label{batches relation}
  U_f^{batch_j} \approx U_f^{batch_k} W, \qquad \text{with}, \qquad W:= \left(U_f^{batch_k}\right)^T\left(U_f^{batch_j}\right)^T.
\end{equation}
Accordingly, all bases found using different batches are close to being linear combinations of each others, hence backing up Observation \ref{obs: batch}.

Since the bases found in different batches are nearly linear combinations of each other, we can determine the final training basis when training with batches in two ways,
\begin{enumerate}
  \item \underline{The naive approach:} We assume that the batches are close enough to being linear combinations and choose any batch basis as the final training basis. To explain why this would work, assume we encoded a sample $X_i$ with a certain batch $k$, but are using the basis from a different batch $j$ for the projection, we can write the truncated latent space as,
  \begin{equation}
    U_f^{batch_j}\left(U_f^{batch_j}\right)^T Y_i^{(k^*)} = U_f^{batch_j}\left(U_f^{batch_j}\right)^TU_f^{batch_k}\alpha^{batch_k}_i,
  \end{equation}
  where we used equation \eqref{trunc projection} for the second part of the equation. Based on equation \eqref{batches relation}, we can re-write the above as,
  \begin{equation}\label{new alpha}
    \begin{aligned}
    U_f^{batch_j}\left(U_f^{batch_j}\right)^T Y_i^{(k^*)} &\approx U_f^{batch_k}WW^T\underbrace{\left(U_f^{batch_k}\right)^TU_f^{batch_k}}_{=I_k}\alpha^{batch_k}_i,\\
    &= U_f^{batch_k}\underbrace{WW^T\alpha^{batch_k}_i}_{:=\alpha^{new}_i}.\\
    \end{aligned}
  \end{equation}
  Equation \eqref{new alpha} shows that no matter which basis $U_f^{batch_j}$ we choose for the projection, the result would be similar to projecting to basis $U_f^{batch_k}$ with a new set of coefficients $\alpha^{new}_i$. Accordingly, when the bases are sufficiently close to being linear combinations of each others, any batch basis can be used for the projection. While this seems to work well empirically, we use a more involved approach throughout the paper to alleviate the convergence and linear independency requirements.
  \item  \underline{The involved approach - fine tuning:} We can find the final training basis by computing a common basis to all the batches bases found during training. This would be done by concatenating the bases found for the batches and performing again a truncated SVD as follows,
  \begin{equation}\nonumber
    U_f\alpha_f = SVD_{k^*}\left(\left[\begin{array}{c|c|c|c}
      U_f^{batch_1} & U_f^{batch_2} & \cdots & U_f^{batch_{D//B}}
    \end{array}\right]\right)
  \end{equation}
  with $D//B$ being the number of batches required to cover the entire data set, note that this approach is more generic. If the bases are linear combinations of each other, the matrix of concatenation above will have a rank of $k^*$, and the basis found will perfectly represent all the batch bases. On the other hand, if the training was not thorough, or if there aren't enough samples that are linearly independent, the differences between the linear combinations of the bases will be truncated by SVD.

  Since the projection onto the final basis found will be slightly different compared to the previous bases during training, the decoder must be informed of the common basis $U_f$. In this case, we propose a \textbf{fine-tuning procedure}, where, after training, the common basis is found as explained above, and the decoder is trained for a few forward passes while projecting onto the new basis to capture any noise caused by the differences between $U_f$ and the batch bases. During fine tuning, it is crucial to fix the encoder to ensure the regularized space found by the encoder when training with the SVD is not modified. While fine-tuning may not be necessary in many cases, we believe it is a good practice to ensure that a common representative basis---and thus a bottleneck---is found for all the samples.
\end{enumerate}
Finally, we summarize the entire training procedure when using batches and a fixed bottleneck as follows:

\begin{itemize} 
  \item Train the encoder/decoder by passing batches to optimize the reconstruction loss with a truncated SVD of rank $k^*$ in the latent space. 
  \item Before the last epoch, pass the data set in batches to the encoder to find its latent space, perform a truncated SVD for each batch, and save the truncated basis $U_f^{batch_i}$ for each batch $i$. 
  \item Concatenate all the bases found for the batches and perform a truncated SVD to find the final training basis $U_f$. 
  \item For the last epoch, fix the encoder, and fine-tune the decoder using the final training basis $U_f$. In this case, the truncated SVD in the latent space is replaced by a projection onto $U_f$ (i.e., left-multiplying the latent space by $U_f U_f^T$). 
  \item For inference, the projection is done onto the final training basis $U_f$, similarly to how it was previously done in Subsection \ref{subsec: RRAE_fixed_whole}.
\end{itemize}

\section{Adaptive Rank Reduction Auteoncoders (aRRAEs)}\label{sec: aRRAE}
The model and training procedure presented above were designed assuming the bottleneck size is known a priori. However, one of the main advantages of RRAEs is that they alleviate the direct dependence of the encoder/decoder architectures on the bottleneck size. Furthermore, as discussed earlier, the singular values determine the importance of the terms in the bottleneck, with these values being sorted in decreasing order based on their significance. A natural extension of RRAEs, therefore, is to develop an adaptive algorithm that can automatically determine the value of $k^*$. In this section, we present an algorithm to achieve this, outlined as follows:
\begin{algorithm}[H]
  \caption{Adaptive Rank Reduction Autoencoder Algorithm}
  \label{alg:adaptive}
  \begin{algorithmic}
    \State \textbf{Input:} Unlabeled Data $X \in \mathbb{R}^{D\times N}$, stagnation criteria.
    \State Initialize $k^*= \min (D, L) $ (or $\min(B, L)$ if bacthed).
    \State Train RRAEs as described in Section \ref{sec:RRAE} until convergence.
    \State Store the final epoch's loss as the \emph{optimal\_loss}.
    \While {True}
      \State $k^*=k^*-1$.
      \State Continue training the RRAE with the new $k^*$ until one of the following,
      
      \If {\emph{optimal\_loss} reached}
        \State Save model at $k^*$.
      \ElsIf {stagnated}
        \State $k^*=k^*+1$.
        \State Load model saved at $k^*$.
        \State Break.

      \EndIf
    \EndWhile
    \If {batches are used}
      \State Find the final training basis $U_f$ as described in Subsection \ref{subsec: RRAE_fixed_batches}.
      \State Fine tune the decoder using $U_f$.
    \Else
      \State Find the final training basis $U_f$ as in Figure \ref{fig:RRAE}.
    \EndIf
    \State \textbf{Output:} Trained RRAE with the optimal bottleneck size, and $U_f \in \mathbb{R}^{L\times k^*}$.
  \end{algorithmic}
\end{algorithm}

The stagnation criteria are details in Appendix \ref{aRRAEs details}. The algorithm is based on the observation that the last singular values are typically the smallest. Therefore, it removes the last singular value (the one of least importance) and redistributes its effect across the more significant singular values. The initial value of $k^*$ should be set to the maximum possible value. But how do we determine the maximum rank $k^*$? The rank of a matrix is always bounded above by its dimensions. Since we perform the SVD on the latent space matrix, which is of shape $L \times D$ (or $L \times B$ if batched), the maximum possible value of $k^*$ is the minimum between either $B$ (batched) or $D$ (unbatched), and $L$.

This algorithm represents the simplest version to demonstrate the concept. Future work will include enhancements such as striding (such as skipping multiple $k^*$ values based on certain criteria) or accepting a tolerance instead of always aiming for the optimal loss. However, we show through both real and synthetic data sets that, in its simplest form, the algorithm is sufficient to find an\footnote{Note that we say “an” optimal value because, in some cases, multiple values can lead to good results.} optimal value of $k^*$, resulting in improved interpolation and better sample generation. Furthermore, we demonstrate that starting with a larger value of $k^*$ and then adaptively reducing it leads to better representation quality and improved model generalization, as the model has access to more information in the beginning of the training process.

\section{Results and Discussion}
In this section we test Rank Reduction Autoencoders both with and without the adaptive algorithm on multiple data sets. We first present the results on a synthetic data set with a highly local behavior. Then we present results on the MNIST, fashion-MNIST, and CelebA data sets and show that RRAEs both with/without the adaptive scheme are stable and provide advantages over other regularizing autoencoders. Further, we perform both a scalability analysis and an ablation study to show the importance of the truncated SVD in the latent space and the capability of RRAEs to learn features when the size of the bottlneck is large.

The purpose of this section is to show experimentally two things,
\begin{itemize}
  \item RRAEs with a fixed bottleneck are able to learn a good representation of the data, even for samples with localized behavior, and others that require large bottlenecks.
  \item The adaptive algorithm is able to find an optimal value of $k^*$ independently of the magnitude of $k^*$. The adaptive algorithm is also useful to get more meaningful latent spaces since the network has access to more information in the beginning of the training process.
\end{itemize}

Throughout the paper, all models except the diabolo Autoencoder share the same architecture. Sparse AEs include an additional sigmoid activation function at the end of the encoder, IRMAE includes $l$ additional linear layers before the latent space, and RRAEs include an SVD\footnote{For machines with higher precision, we use a \emph{Stable SVD} to avoid gradient explosions, details are available in Appendix \ref{stable_SVD}.} in the latent space. For both RRAEs without the adaptive algorithm, and Diabolo AEs, a bottleneck size has to be chosen a priori. For all examples below, we first train the adaptative RRAE to find a good value of $k^*$, then use the same $k^*$ to define the architecture of both the Diabolo AE and the RRAE with a fixed bottleneck. More details about model architectures and training parameters can be found in Appendix \ref{appendix_train}.
\subsection{Synthetic Data Set}\label{subsec: synthetic}
We begin by testing Rank Reduction Autoencoders (RRAEs) and comparing them to other methods on a synthetic data set with localized behavior and a predefined bottleneck size. For this experiment, we selected a local function: a Gaussian bump in a 2D space. A $64 \times 64$ 2D grid was defined, with the magnitude and variance of the Gaussian bump fixed at 1 and 0.2, respectively. The data set was generated by varying two parameters: the positions of the bump along both the horizontal and vertical axes. In other words, the data set consists of Gaussian bumps with consistent magnitude and variance, but displaced along both axes of the domain. We generated 600 samples, with bump positions sampled on an equidistant grid. The test set consisted of randomly generated points within the range of the training data.

On this data set, we trained RRAEs with a fixed bottleneck size of 2, RRAEs using the adaptive algorithm, as well as Diabolo, Sparse, Contractive, and Implicit Rank Minimizing Autoencoders. We evaluated the performance of the models based on their reconstruction errors on both the training and test sets, as well as their ability to generate new samples. To assess this, we generated samples in two ways:

\begin{enumerate} 
\item \underline{Interpolation:} If the latent space is learned correctly, a linear interpolation between two samples in the latent space should produce meaningful samples in the data space. To test this, we randomly selected 50 pairs of samples and linearly interpolated between them, generating five intermediate images for each pair. 
\item \underline{Random Sampling:} We randomly sampled 250 points in the latent space, using a normal distribution with the mean and covariance of the training latent space. 
\end{enumerate}

\begin{figure}[!b]
  \centering
  \includegraphics[width=0.8\textwidth, trim=0 40 0 40, clip]{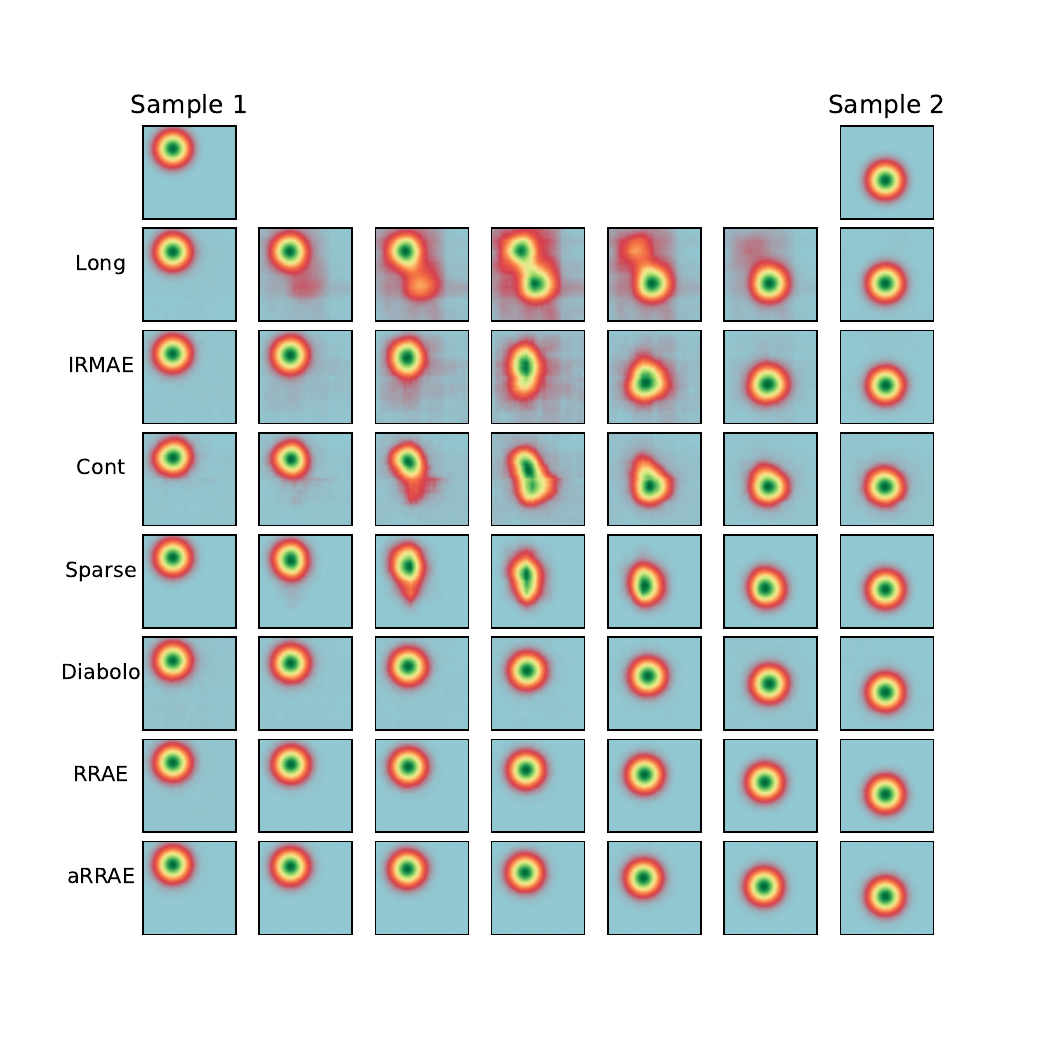}
  \caption{Interpolation for different architectures on the 2D gaussian problem.}
  \label{fig:interpolation_gauss}
\end{figure}

\begin{figure}[!t]
  \centering
  \includegraphics[width=0.8\textwidth, trim=0 40 0 40, clip]{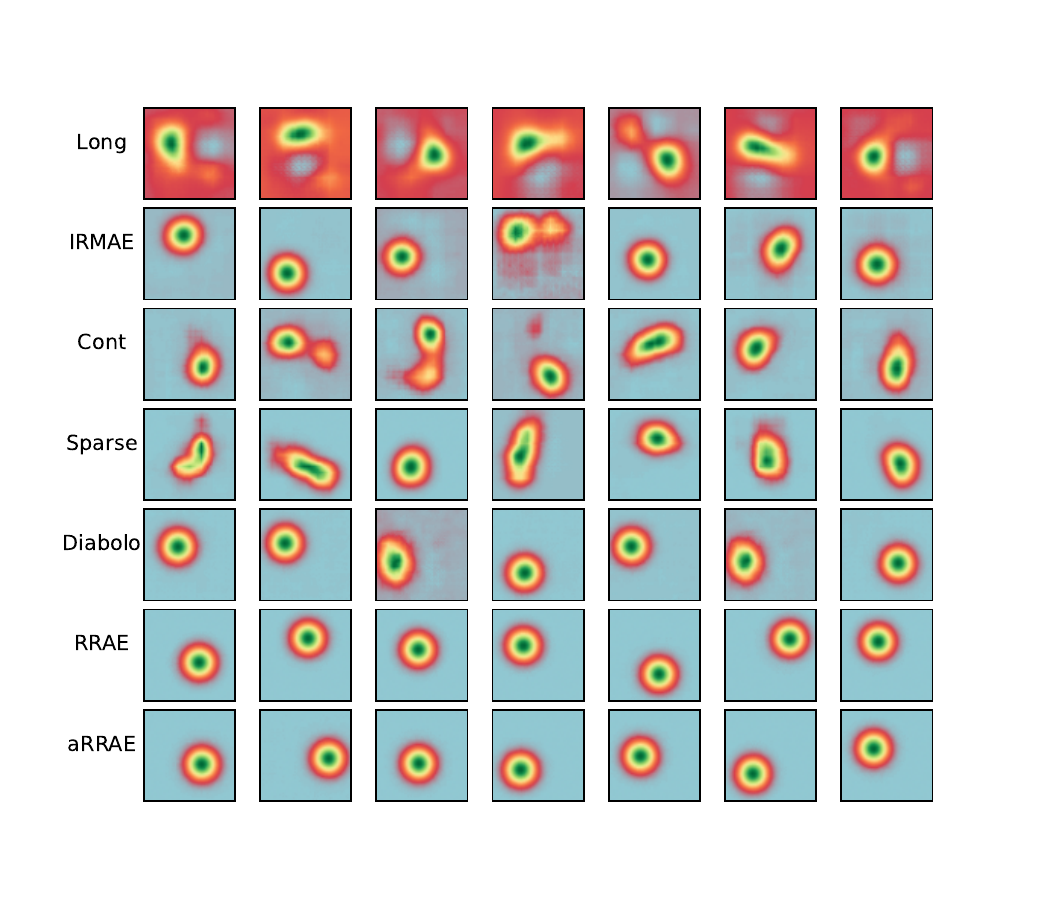}
  \caption{Random sampling for different architectures on the 2D gaussian problem.}
  \label{fig:random_sampling}
\end{figure}

Examples of interpolation and random sampling for the different models can be found in Figures \ref{fig:interpolation_gauss} and \ref{fig:random_sampling}, respectively. The figures demonstrate that RRAEs and aRRAEs can generate samples that closely resemble a Gaussian, while other models, including the Diabolo AE, struggle to do so. The good performance of both RRAEs and aRRAEs is explained by both the bottleneck and the regularization imposed by the truncated SVD during training.

To quantify the results, the reconstruction error was calculated as the percentage difference in the $L_2$-norm between the original sample and its reconstruction. The generated images were evaluated by fitting a Gaussian bump with the correct amplitude and variance to each generated sample. We then computed the percentage difference in the $L_2$-norm between the fitted bump and the original bump. The results of this evaluation are presented in Table \ref{fig:table_shift_sin}.
\begin{table}[!h]
  \centering
  \begin{tabular}{ccccc}
      \toprule
      Model         & Train Error                      & Test error                     & Interpolation   & Random   \\
      \midrule
      Long    &    1.95                         & 2.07                          & 26.46 $\pm$ 15.2          & 63.29 $\pm$ 12.39         \\
      IRMAE & 3.66                   & 3.78                  & 20.02 $\pm$ 18.68 & 39.17 $\pm$ 16.87 \\
      Cont & 10.76                   & 12.04                  & 27.51 $\pm$ 17.4 & 57.81 $\pm$ 16.13 \\
      Sparse     & \textbf{0.92}                            & \textbf{1.21}                           & 17.29 $\pm$ 19.96          & 50.94 $\pm$ 21.15          \\
      \midrule
      Diablo     & 2.66                            & 2.45                           & 1.93 $\pm$ 0.34          & 4.28 $\pm$ 4.79          \\
      RRAE (fixed)  & \textbf{0.98}                          &  \textbf{1.14}                          & 1.03 $\pm$ 0.39          & 1.78 $\pm$ 1.2          \\
      aRRAE         & 0.99                            & 1.27                           & \textbf{0.982 $\pm$ 0.08}          & \textbf{1.26 $\pm$ 0.43}          \\
      \bottomrule
  \end{tabular}
  \caption{Relative error (in \%) for different architectures on the 2D gaussian problem. The error represents the difference between the original bump and the reconstruction (columns 1 and 2), and the difference between the fitted bump and the generated one (columns 3 and 4). Mean and standard deviations for generated samples are reported over 250 samples. A horizontal line in the middle splits AEs that don't enforce a bottleneck (top) and those who do (bottom).}
  \label{fig:table_shift_sin}
\end{table}

The results show that RRAEs with a fixed bottleneck are able to learn a strong representation of the data. Whether by comparing reconstruction errors or assessing the quality of the latent spaces, RRAEs outperform a Vanilla Diablo Autoencoder with a fixed bottleneck, underscoring the importance of the regularization imposed on the latent space bottleneck. On the other hand, the adaptive algorithm successfully determined an optimal value of $k^* = 2$, which matches the size of our parametric space, while also producing the most interpretable bottleneck (this is represented by generating the best Gaussian bumps when sampling from the latent space). It’s important to note that Autoencoders that do not enforce a bottleneck can achieve lower errors on the training set. This is expected, as the bottleneck introduces a loss of information in the network. However, the bottleneck is also one of the main reasons for having a more interpretable latent space. 

\begin{figure}[!t]
  \centering
  \includegraphics[width=1\textwidth, trim=0 3 0 40, clip]{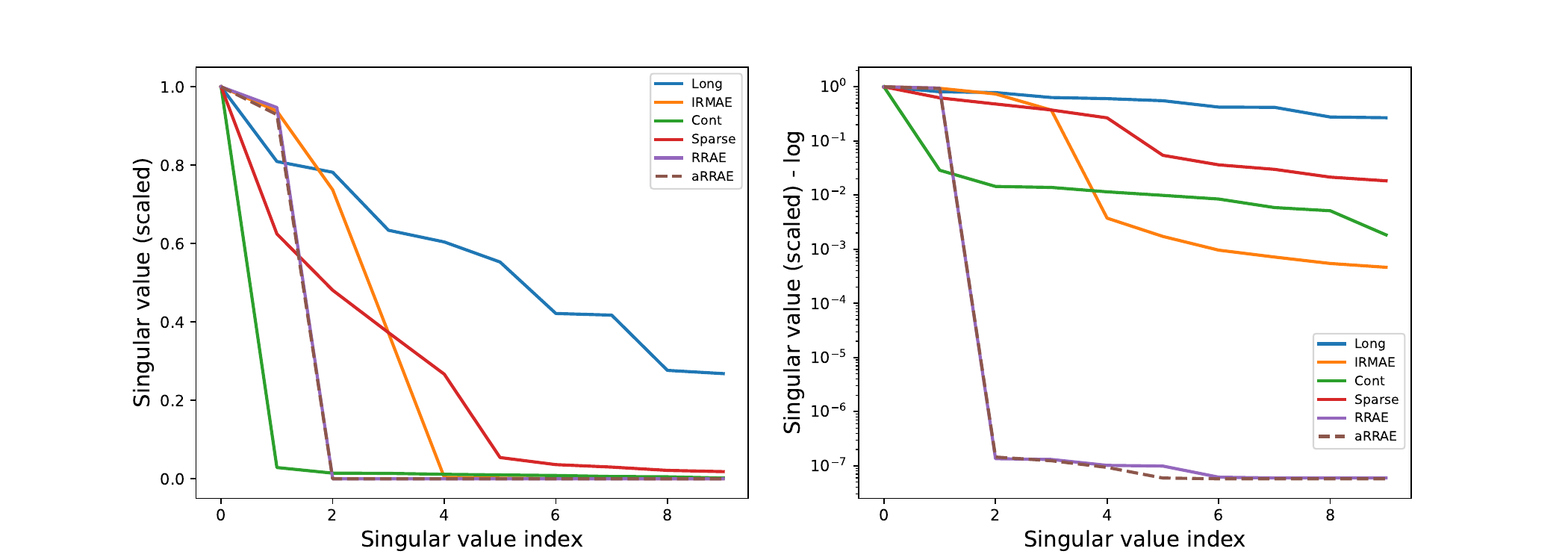}
  \caption{Singular values of the latent space matrices for different architectures on the 2D gaussian problem. Log-scale to the right.}
  \label{fig:singular_values}
\end{figure}

To better understand the good performance of RRAEs, we plot the singular values of the training latent space matrix, scaled by the inverse of the largest singular value, in Figure \ref{fig:singular_values}. Both RRAEs and aRRAEs learn a latent space with rank $k^*$, having exactly $k^*$ non-zero singular values. In contrast, the other Autoencoder architectures, as shown on the log scale to the right, do not force the remaining singular values to approach zero. Non-zero singular values suggest that there isn't a bottleneck, causing less meaningful latent spaces.

\subsection{Real World Data Sets}
In this subsection, we present the results comparing the performance of both Rank Reduction Autoencoders (RRAEs) and adaptive RRAEs (aRRAEs) with other models on typical real-world data sets used in machine learning. We selected three data sets: MNIST (\citealp{deng2012mnist}), Fashion MNIST (\citealp{xiao2017fashionmnistnovelimagedataset}), and CelebA (\citealp{liu2015faceattributes}), each with a different bottleneck size and varying levels of complexity. As in Subsection \ref{subsec: synthetic}, we evaluate the performance of the models by examining the significance of their latent spaces. For each data set, this is assessed through two approaches:

\begin{enumerate} 
\item \underline{Image Generation via Interpolation:} Although autoencoders are not expected to interpolate between any two images in the data set, if the latent space is sufficiently well-structured, interpolation between multiple image pairs should yield reasonable results. 
\item \underline{Image Generation via Random Sampling:} Similar to the procedure described for the synthetic data set, we assess the ability of the models to generate realistic samples by randomly sampling points from their latent spaces.
\end{enumerate}

\begin{figure}
  \centering
  \includegraphics[width=1\textwidth, trim=40 550 0 0, clip]{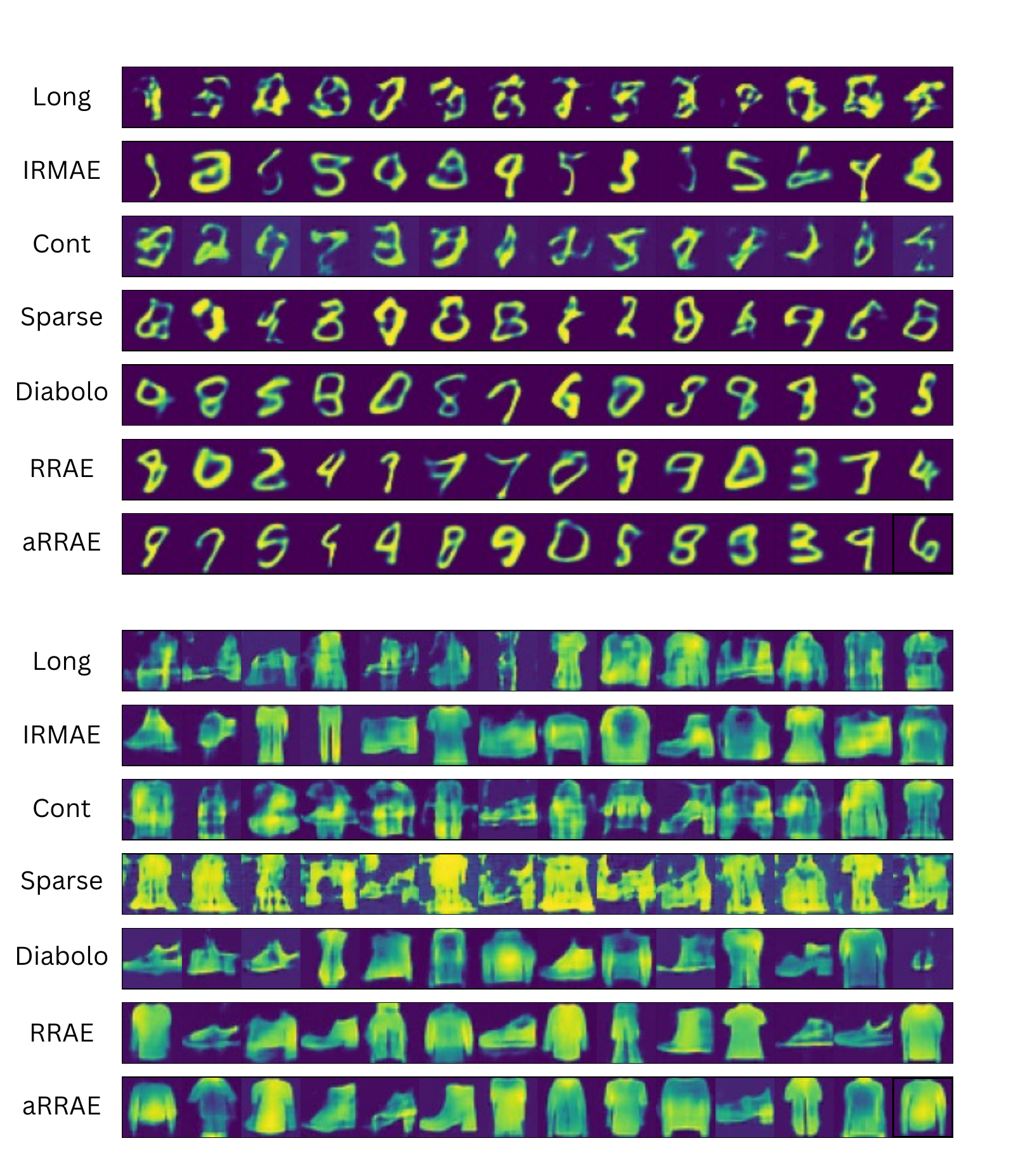}
  \caption{Example of randomly sampled MNIST/ fashion MNIST images using all Autoencoders presented in the paper.}
  \label{fig:random_both_mnists}
\end{figure}

\begin{figure}
  \centering
  \includegraphics[width=1\textwidth, trim=40 450 0 0, clip]{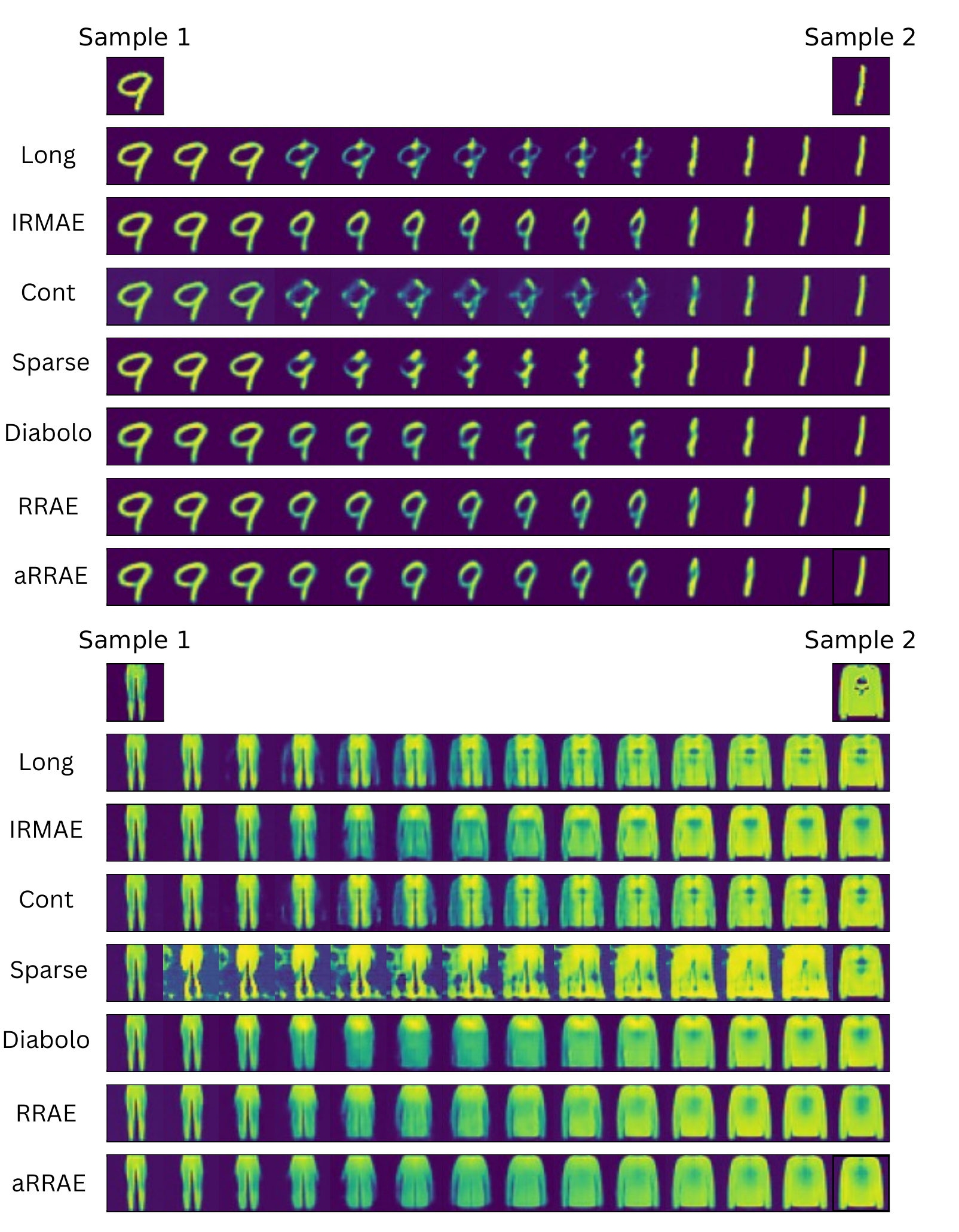}
  \caption{Example of interpolation between two MNIST/fashion MNIST samples using all Autoencoder architectures presented in the paper.}
  \label{fig:interpolation_both_mnists}
\end{figure}

In this case, the quality of the generated images is evaluated by fine-tuning an Inception-V3 model on each data set, removing its head, and using the output features to compute the Fréchet Inception Distance (FID) score between the generated images and the real ones. Some newly generated images by randomly sampling from the latent spaces of each model can be found in Figures \ref{fig:random_celeba} and \ref{fig:random_both_mnists}. Additionally, interpolated sample images are shown in Figures \ref{fig:interpolation_celeba} and \ref{fig:interpolation_both_mnists}. For CelebA, only interpolated images generated by aRRAEs are shown, as the other architectures produce blurry images, as evidenced by the random samples shown in Figure \ref{fig:random_celeba}.
\begin{figure}
  \centering
  \includesvg[width=0.8\textwidth]{interp_Celeba_aRRAE_1.svg}
  \caption{Example of interpolation between different samples using aRRAEs. Other models are not presented since their reconstruction is blurry. The picture to the right in the first row is the real image flipped along the vertical axis.}
  \label{fig:interpolation_celeba}
\end{figure}
\begin{figure}
  \centering
  \includesvg[width=0.9\textwidth]{random_1_1.svg}
  \caption{Example of randomly sampled CelebA images using all Autoencoders presented in the paper.}
  \label{fig:random_celeba}
\end{figure}
 The FID scores of the generated images, as well as the reconstruction errors, and the classification accuracies can be found in Table \ref{fig:table_real_world}.
\begin{table}[!h]
  \centering
  
  \begin{tabular}{ccccccc}
        \toprule
        {}            & \multicolumn{2}{c}{MNIST} & \multicolumn{2}{c}{Fashion-MNIST} & \multicolumn{2}{c}{CelebA}                               \\
        \cmidrule(lr){2-7}                                                                                               \\
        \addlinespace[-0.2cm]
        Model         & Interp.                      & Random                     & Interp.   & Random & Interp. & Random   \\
        \midrule
        Long    &    10.16                         & 86.5                          & 35.21          & 67.13 &   16.53 &   17.23     \\
      IRMAE & 8.09                   & 42.58                  & 23.2 & 43.62 & 15.79 & 15.47 \\
      Cont & 30.5                   & 90.5                  & 27.51  & 55.32  & 16.74 & 17.36 \\
      Sparse     & \textbf{6.1}                          & 57.67                          & 61.32          & 87.2    & 16.91 & 16.57      \\
      \midrule
      Diablo     & 8.92                            & 44.5                           & 23.6          & 48.29  & 15.76 & 15.75        \\
      RRAE (fixed)  & 7.45                          &  41.5                          & 23.41          & 40.6   & 14.89 & 15.07       \\
      aRRAE         & 6.6                            & \textbf{41.42}                          &  \textbf{22.31}          & \textbf{40.21} & \textbf{5.74} & \textbf{5.91}          \\
        \bottomrule
    \end{tabular}
    \caption{Table documenting FID values for 20000 images generated by Interpolation/Random sampling. Standard deviations were not given as they were all of a small order.}
\label{fig:table_real_world}

\end{table}

The results presented above highlight the advantages of using either Rank Reduction Autoencoders (RRAEs) or adaptive RRAEs (aRRAEs) for learning features from unlabeled data. It is important to note that, due to the size of the data sets, Long Autoencoders (Long AEs) also produce acceptable interpolated and generated images. This is primarily because the abundance of data points helps mitigate the ``holes'' in the latent space, as discussed earlier. While aRRAEs demonstrate superior performance across all data sets, their advantage is particularly pronounced with the CelebA data set.

\begin{figure}
  \centering
  \includegraphics[width=1\textwidth, trim=0 3 0 40, clip]{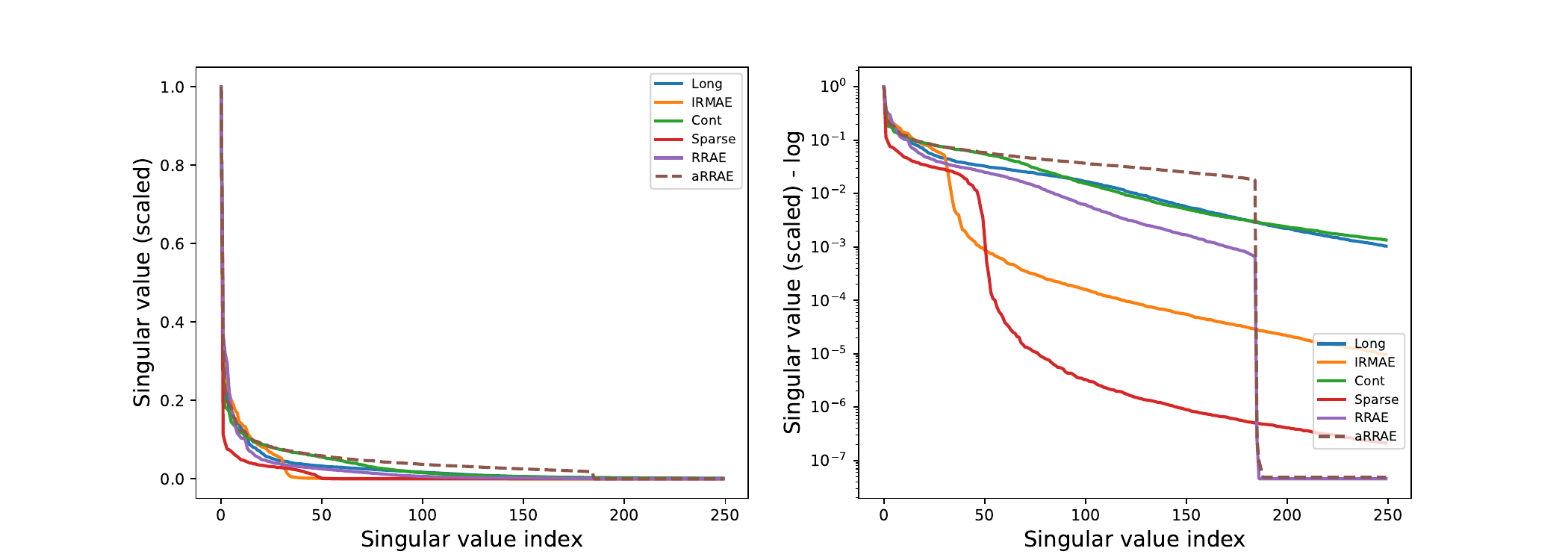}
  \caption{Singular values of the latent space matrices for different architectures on the CelebA problem. Log-scale to the right.}
  \label{fig:singular_values_celeba}
\end{figure}

To further investigate why aRRAEs outperformed the other architectures, we examine the singular values of the latent space, both on regular and logarithmic scales, as shown in Figure \ref{fig:singular_values_celeba}. As observed, all architectures tend to emphasize only a few dominant singular values, which explains why their interpolation and random generation results are relatively satisfactory. However, the log-scale plot reveals that only RRAEs and aRRAEs maintain exactly 186 significant features, corresponding to the size of their bottleneck.

This bottleneck structure explains why RRAEs perform better than the other models. However, aRRAEs provide a further advantage, enabling them to achieve an even lower FID score. Notice that the singular values for aRRAEs are significantly larger than those of other models, sharply dropping to zero at index 187. The adaptive algorithm allows the model to effectively inject the impact of the features into the compressed representation, guiding the latent space to maximize the amount of information retained. This is a significant advantage compared to models that start with a pre-compressed representation, where the network must learn to adapt to a fixed bottleneck from the beginning.

\subsection{Ablation Study}
The results presented in the previous sections demonstrate that both RRAEs and aRRAEs exhibit more meaningful latent spaces compared to both Vanilla AEs with a long latent space and Diabolo AEs with a fixed bottleneck. In this section, we conduct an ablation study to empirically show that the use of SVD is essential, and that learning a truncated basis without the SVD is insufficient. We present three alternative approaches to compare with RRAEs:

\begin{itemize} 
\item \underline{Fix-Basis AE:} In this approach, we train a standard autoencoder (AE) with a long latent space, using a pre-determined orthonormal basis $U_f$ that is fixed throughout training. The primary purpose of this comparison is to demonstrate that a long latent space, when simply projected onto a reduced basis, is not sufficient. This also underscores the importance of the regularization imposed by the SVD on the latent space. 
\item \underline{Learn-Basis AE:} In this model, we train an AE with a long latent space and a trainable basis $U_f$ that is learned during training. This model removes the assumption made in the previous approach, where the basis was fixed. Here, the basis is simply a trainable matrix and does not rely on the output of the encoder, unlike the SVD. 
\item \underline{Weak RRAEs:} This model involves training a standard AE with a long latent space, while enforcing the latent space matrix to be low rank by adding the difference between the latent matrix $Y$ and its truncated SVD to the loss function. This imposes a weaker form of the same properties as RRAEs, as the low-rank constraint is applied during training through the loss function rather than through the SVD. 
\end{itemize}

\begin{figure}[!h]
  \centering
  \includegraphics[width=1\textwidth, trim=0 120 0 140, clip]{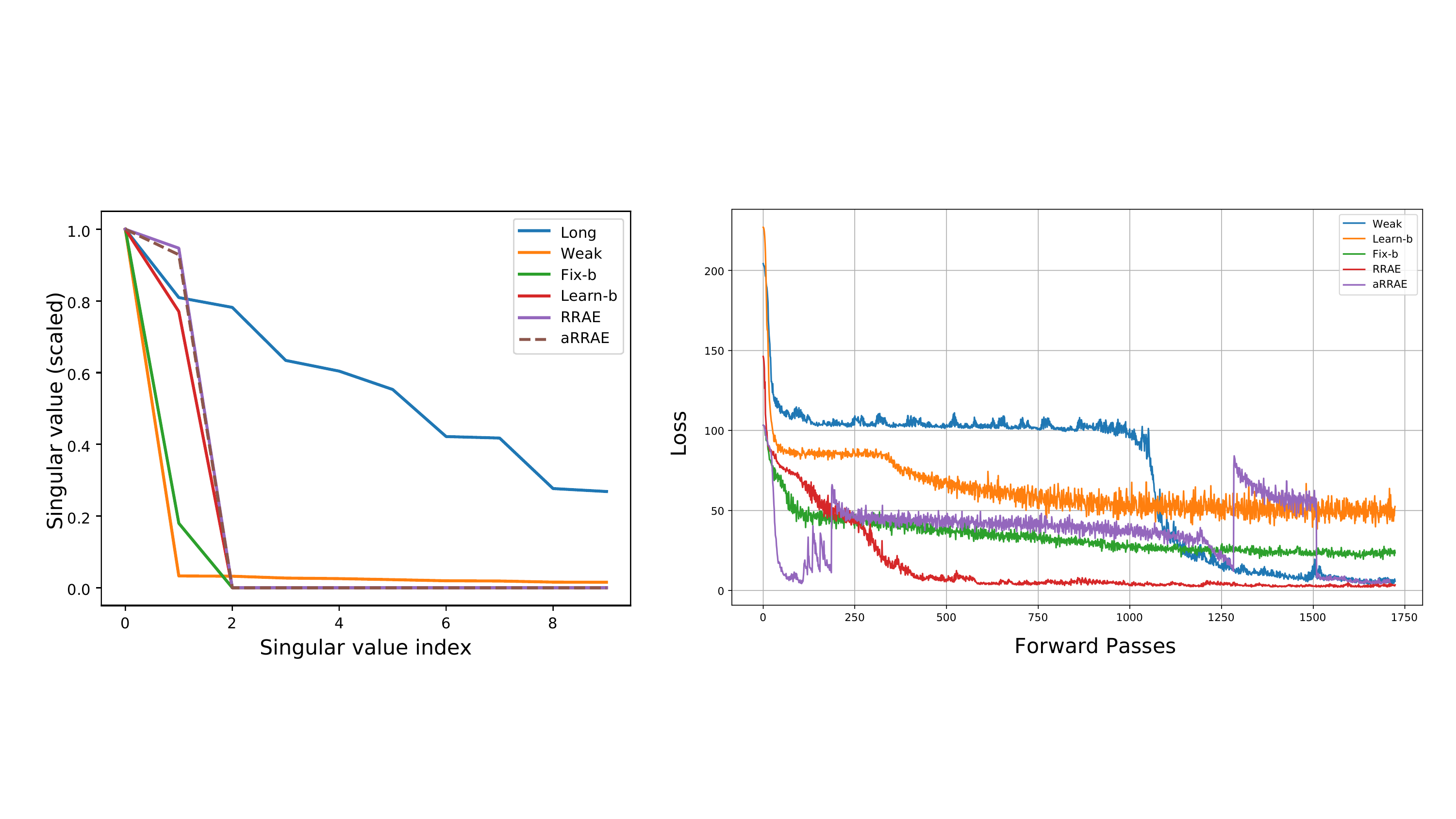}
  \caption{Singular values for different models (left), and training loss evolution (right).}
  \label{fig:sv_and_loss}
\end{figure}

We train all three models outlined above on the synthetic data set introduced in Subsection \ref{subsec: synthetic} and compare their performance to RRAEs. The singular values and loss evaluation during training can be found in Figure \ref{fig:sv_and_loss}.

As can be seen in the figure, the Weak AE does not find a bottleneck since its small singular values are nonzero. On the other hand, while Fix-basis and Learn-basis AEs are designed to have a bottleneck, their convergence is worse as can be seen by the evolution of the loss. Note that both RRAEs and aRRAEs converge fast during training. The jumps in the loss value for aRRAEs is expected when a singular value is removed by the adaptive algorithm. Notice how around the 1250-th forward pass, aRRAEs tried to train with one mode (hence the increase in error and stagnation), but then the algorithm chose to go back to two modes and finish the training there. The speed of convergence of aRRAEs depends on the algorithm chosen. Fine tuning the algorithm further to make it faster is left for future research. The results show empirically the significance of the SVD inside the forward pass and highlight all of the advantages presented in Section \ref{sec:RRAE}.
An example of interpolated images can be found in Figure \ref{fig:interp_ablation}, the errors can be found in Appendix \ref{ablation_err}.

\begin{figure}[!h]
  \centering
  \includegraphics[width=0.7\textwidth, trim=0 0 0 0, clip]{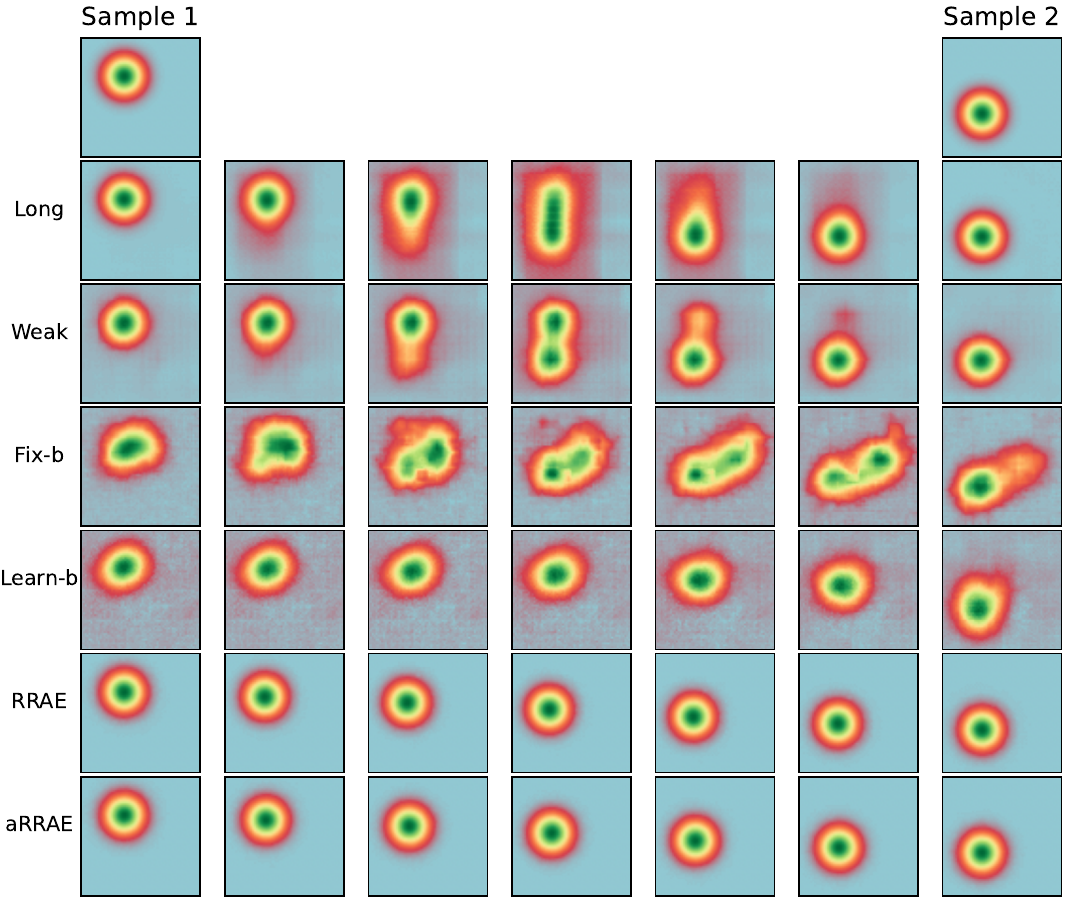}
  \caption{An example of interpolation performed for the ablation study.}
  \label{fig:interp_ablation}
\end{figure}

\begin{figure}[!t]
  \centering
  \includegraphics[width=0.7\textwidth, trim=0 50 0 30, clip]{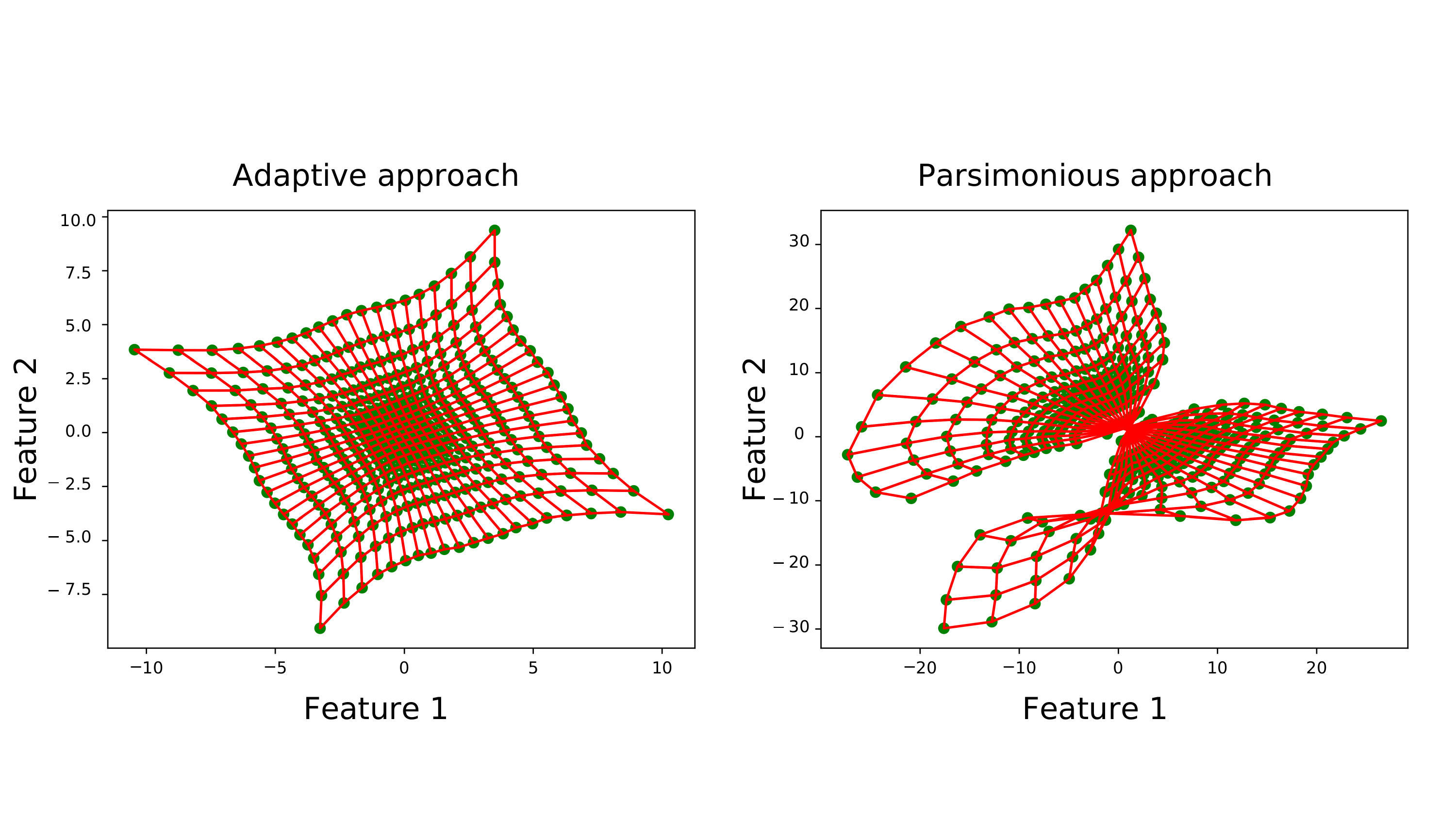}
  \caption{Second latent feature vs the first one for both adaptive algorithms. The red lines are drawn according to neighbors in the original parametric space. Accordingly, closed polygons formed by red lines mean that neighboring is kept in the latent space.}
  \label{fig:coeffs_pars}
\end{figure}

We continue further in our ablation study to test if starting with a large number of singular values in the adaptive algorithm is beneficial compared to starting with one singular value and increasing them during training (i.e., a parsimonious approach). To compare these ideas, we trained aRRAEs with both algorithms on the synthetic data set. While both algorithms were able to find the optimal value of $k^* = 2$, the latent space using the parsimonious algorithm is less interpretable wether for interpolation, or random generation (again, the errors can be found in Appendix \ref{ablation_err}). This is mainly due to the fact that when training with fewer singular values than needed, the RRAE tries to find the best approximation possible without expecting other singular values to be added. Accordingly, it is easier to converge to a less meaningful singular value, and have a hard time getting out of the local minimum when a new singular value is added. To illustrate the limitation of the parsimonious approach, we plot the first latent feature against the second one when using both algorithms for the gaussian problem in Figure \ref{fig:coeffs_pars}. The figure illustrates how the algorithm presented in the paper conserves the neighboring of the original space. On the other hand, the parsimonious approach learns a less meaningful latent space, especially near the center, where many red lines intersect at a point and interpolation/generation become problematic. 
\subsection{Scalability of RRAEs}
The results above showcased how RRAEs can be used with a bottleneck of average size. Yet, the largest value of $k^*$ found by the algorithm was $186$ for the CelebA data set. However, since the bottleneck is determined by the singular values, one might wonder wether RRAEs are capable to learn compressed features when the bottleneck is of larger size. Accordingly, we conducted on experiment where RRAEs where trained on the CelebA data set with multiple fixed values of $k^*$. The reconstruction for each value of $k^*$ can be found in Figure \ref{fig:scalability}. With the stable SVD presented in Appendix \ref{stable_SVD}, RRAEs can learn bottlenecks of larger sizes reaching a size of $2500$. In this case we stopped increasing $k^*$ after $2500$ since the reconstructions started getting very close to the original picture. The reconstruction errors for different $k^*$ values can be found in Table \ref{fig:table_scal}.

\begin{figure}[!h]
  \centering
  \includegraphics[width=0.6\textwidth, trim=0 0 0 0, clip]{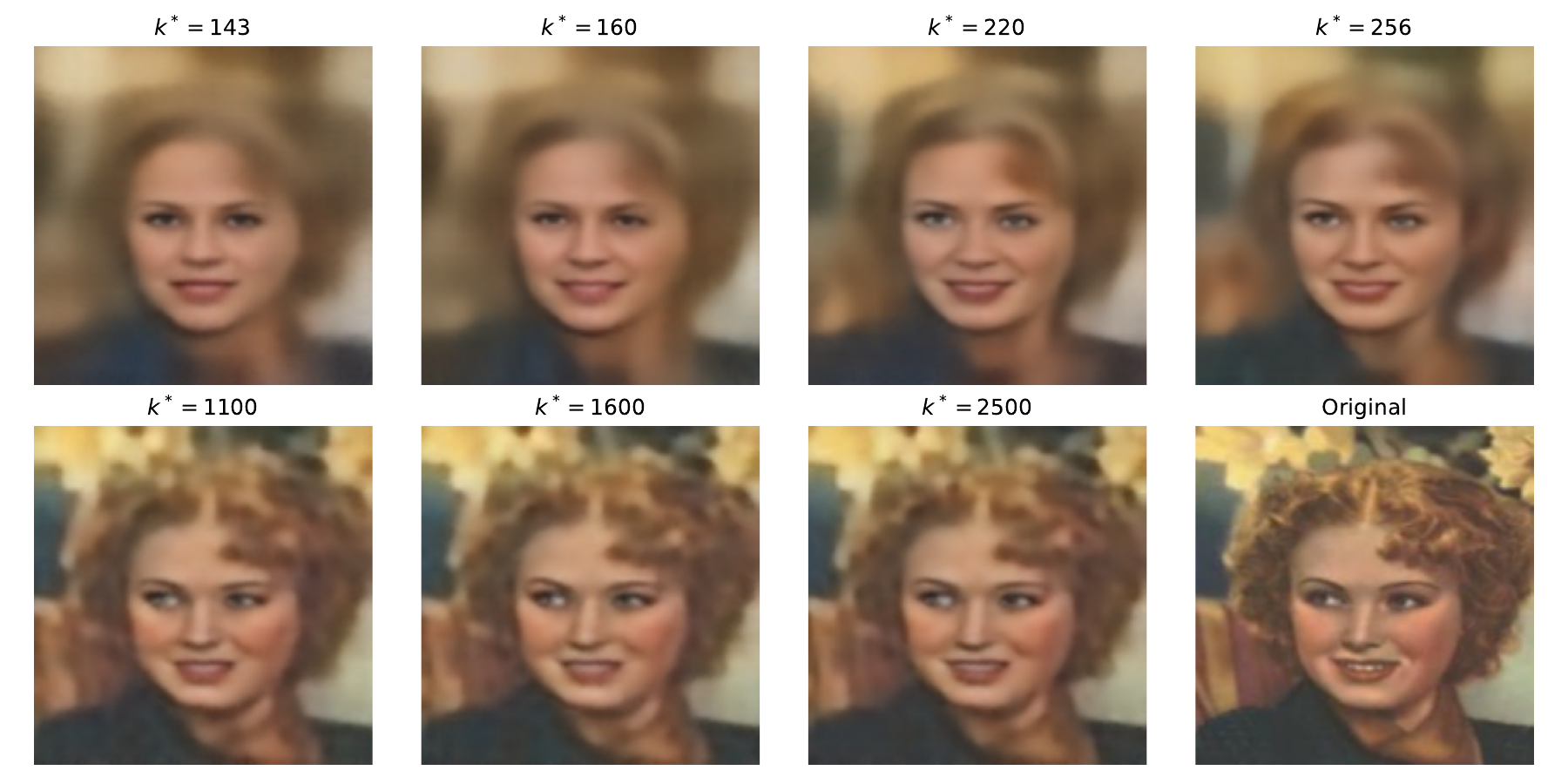}
  \caption{Reconstructed images using RRAEs with different $k^*$ values.}
  \label{fig:scalability}
\end{figure}

\begin{table}[!h]
  \centering
  \begin{tabular}{cccccccc}
      \toprule
      $k^*$         & 143                      & 160                     & 220 & 256 & 1100 & 1600 & 2500   \\
      \midrule
      Train Error & 15.91                   & 15.04                  & 13.7 & 13.11 &  8.88 & 8.42 & 8.1 \\
      Test Error & 16.04                   & 15.18                & 13.84 & 13.27 & 9.04 & 8.6 & 8.3 \\
      \bottomrule
  \end{tabular}
   \caption{Reconstruction Errors (in \%) for RRAEs with different $k^*$ values over the CelebA data set.}
  \label{fig:table_scal}
\end{table}

\subsection{Hyperparameter dependence}\label{HP}
All the results presented above were obtained using the optimal hyperparameters identified during training. However, as previously mentioned, Sparse, Contractive, and Implicit Rank Minimizing Autoencoders each require tuning at least one hyperparameter. For all these architectures, training had to be repeated with various combinations of hyperparameters to determine the optimal values. The hyperparameters for each model and data set can be found in Table \ref{fig:table_hp}. The values listed in the table highlight the significant variation in hyperparameter choices across different problems. 

\begin{table}[!h]
  \centering
  \begin{tabular}{cccc}
      \toprule
      Model         & MNIST                      & Fashion-MNIST                     & CelebA   \\
      \midrule
      IRMAE & $l=8$                   & $l=4$                  & $l=4$ \\
      Cont & $\beta=1000$                   & $\beta=0.1$                 & $\beta=0.00001$ \\
      Sparse     &  $\beta = 1, \, \rho = 0.0001$          &
      $\beta =1\mathrm{e}{-5} , \, \rho = 0.1$ & $\beta = 100, \, \rho = 0.001$          \\
      \bottomrule
  \end{tabular}
  \caption{Best hyperparameters chosen for Implicit Rank Minimizing, Sparse, and Contractive Autoencoders.}
  \label{fig:table_hp}
\end{table}

\begin{figure}[!b]
  \centering
  \includegraphics[width=1\textwidth, trim=0 0 0 0, clip]{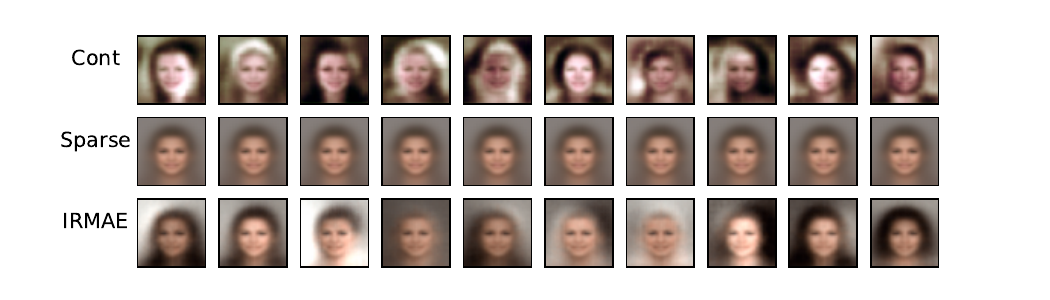}
  \caption{Example of generated sample with wrong hyperparameters. Contractive AE ($\beta = 0.001$), Sparse AE ($\beta = 1, \, \rho = 0.0001$), IRMAE ($l = 8$).}
  \label{fig:images_wrng_hp}
\end{figure}

On the other hand, an incorrect choice of hyperparameters can lead to poor results. For example, we present some generated samples from the Sparse, Contractive, and Implicit Rank Minimizing Autoencoders with suboptimal hyperparameter choices in Figure \ref{fig:images_wrng_hp}. Despite the hyperparameters being selected logically and remaining relatively close to those that produced the best results in other cases, the quality of the generated images is still poor. In contrast, both RRAEs and aRRAEs do not require any additional hyperparameters, making them more reliable and easier to implement in practice.

\section{Conclusion}
This paper introduces a novel class of deterministic autoencoders, Rank Reduction Autoencoders (RRAEs), which regularize their latent spaces through a truncated singular value decomposition (SVD). By incorporating a truncated SVD during the forward pass, RRAEs outperform standard bottleneck architectures and offer more meaningful latent spaces compared to Autoencoders with long latent spaces. Crucially, RRAEs eliminate the dependency of the encoder/decoder architecture on a fixed bottleneck size, as the bottleneck dimension is instead determined by the number of significant singular values in the latent space. Additionally, we present an adaptive version of RRAEs (aRRAEs) that enables the automatic determination of the bottleneck size during training. Experimental results demonstrate that both RRAEs and aRRAEs are stable, scalable, and do not require any additional hyperparameter tuning. The latent spaces learned by these models facilitate meaningful operations on the data manifold, such as interpolation and random sampling, making RRAEs a superior choice for various applications compared to other autoencoder architectures.

\acks{We would like to thank SKF Magnetics Mechatronics for their support in both research and finances. We also thank Google TRC program for giving us access to the required computing power.}


\newpage

\appendix

\section{Gradients and the \emph{Stable SVD}}\label{stable_SVD}
While the SVD offers many regularizing advantages which contribute to the stability of RRAE training, this section presents the gradient of the SVD and demonstrates that the gradients can explode under certain extreme conditions when machines with high precision are used. We then propose a modification to the SVD gradient to numerically prevent gradient explosion. This modified SVD is referred to as the \emph{Stable SVD} and is used in the training of RRAEs.

In automatic differentiation, gradients of functions are defined by their Jacobian-Vector Product (JVP), which enables efficient backpropagation (\citealp{jax_jvp}). The implementation of the JVP for the Stable SVD in JAX is presented below.

\begin{code}
\begin{minted}
  [
  frame=lines,
  framesep=2mm,
  baselinestretch=1.2,
  fontsize=\footnotesize,
  linenos
  ]
  {python}
  @stable_SVD.defjvp
  def _svd_jvp_rule(primals, tangents):
      (A,) = primals
      (dA,) = tangents
      U, s, Vt = jnp.linalg.svd(A, full_matrices=False)

      s = (s / s[0] * 100 >= 1e-9) * s      -->  Added line

      Ut, V = _H(U), _H(Vt)
      s_dim = s[..., None, :]
      dS = Ut @ dA @ V
      ds = _extract_diagonal(dS.real)

      s_diffs = (s_dim + _T(s_dim)) * (s_dim - _T(s_dim))
      s_diffs_zeros = (jnp.abs(s_diffs) <= 1e-9).astype(s_diffs.dtype) --> Changed (was == 0)
      F = 1 / (s_diffs + s_diffs_zeros) - s_diffs_zeros
      dSS = s_dim.astype(A.dtype) * dS
      SdS = _T(s_dim.astype(A.dtype)) * dS

      s_zeros = (s == 0).astype(s.dtype)
      s_inv = 1 / (s + s_zeros) - s_zeros
      s_inv_mat = _construct_diagonal(s_inv)
      dUdV_diag = 0.5 (dS - _H(dS)) * s_inv_mat.astype(A.dtype)
      dU = U @ (F.astype(A.dtype) * (dSS + _H(dSS)) + dUdV_diag)
      dV = V @ (F.astype(A.dtype) * (SdS + _H(SdS)))

      m, n = A.shape[-2:]
      if m > n:
          dAV = dA @ V
          dU = dU + (dAV - U @ (Ut @ dAV)) * s_inv.astype(A.dtype)
      if n > m:
          dAHU = _H(dA) @ U
          dV = dV + (dAHU - V @ (Vt @ dAHU)) * s_inv.astype(A.dtype)

      return (U, s, Vt), (dU, ds, _H(dV))
\end{minted}
\captionof{listing}{Implementation of the jvp for the Stable SVD in JAX. In the code, $\_H$, $\_T$ represent the complex conjugate and the transpose respectively. The changes made to ensure the stability of the SVD stable are in line 7 and 15.}
\label{code:stable_SVD}
\end{code}

As shown in source code \ref{code:stable_SVD}, the gradients of the SVD depend on $F$, which consists of the following,
\begin{equation}\nonumber
    F_{i,j} = \begin{cases}
      \frac{1}{s_i^2 - s_j^2} & \text{if } s_i \neq s_j,\\
      0 & \text{if } s_i = s_j,
    \end{cases}
\end{equation}
with $s_i$ being the $i-$th singular value sorted in decreasing order.

While the case $s_i = s_j$ is handled by setting $F$ to zero, gradients can still explode if the denominator becomes too small. This can occur for two reasons: either the two singular values are very close to each other, or both singular values are very small. We address both of these issues as follows:

\begin{enumerate} 
  \item \underline{Two singular values being close to each other:} To avoid exploding gradients in this case, we modify the definition of $F$ as follows: 
  \begin{equation} \nonumber
    F_{i,j} = \begin{cases} 
      \frac{1}{s_i^2 - s_j^2} & \text{if } |s_i - s_j| > \epsilon,\\
      0 & \text{if } |s_i - s_j| \leq 1e{-9}. 
    \end{cases} 
  \end{equation} 
  Our experimental results show that having a denominator of $1e{-9}$ is sufficient to avoid gradient explosion.

  \item \underline{Both singular values are very small:} To avoid both singular values becoming very small, line 7 is added in the code above. Specifically, we set any singular values smaller than $1e{-9}$ to exactly zero. This ensures that there are no singular values strictly between $0$ and $1e{-9}$, and thus the difference between any two singular values is always greater than $1e{-9}$. Note that having multiple singular values equal to zero is not problematic, as gradient explosions are avoided by the definition of $F$ above.

\end{enumerate}

Note that the value $1e{-9}$ was chosen based on empirical testing and could be tuned for different applications. In general, default thresholds in SVD algorithms automatically truncate singular values to $1e{-6}$, so the above code is not necessary. However, when running the SVD on machines with higher precision (such as TPUs), the default values of the SVD algorithm can be set smaller, making the above code necessary to avoid gradient explosions.

\section{Error for the Ablation Study}\label{ablation_err}
In this section we present the documented errors for the ablation study. The errors (in \%) for the generated gaussian curves can found in the following table.
\begin{table}[!h]
  \centering
  \begin{tabular}{lccccccc}
      \toprule
      Gen Type & Long & Weak & Fix-b & Learn-b & pars-RRAE & RRAE & aRRAE \\
      \midrule
      Interpolation   & 26.46 & 61.57 & 55.18 & 43.79 & 20.41 & 1.03 & 0.982 \\
      Rand. Sampling & 63.29 & 60.5  & 54.87 & 71.03 & 30.21 & 1.78 & 1.26  \\
      \bottomrule
\end{tabular}
\caption{Relative error (in \%) for models tested for the ablation study on the 2D gaussian problem.}
\end{table}
In the table above, pars-RRAE reffers to an RRAE with a parsimonious adaptive algorithm (start with one singular value and add more singular values when the loss stagnates).

\section{Model Architecture and Training}\label{appendix_train}
Throughout the paper, we use the same algorithm to create the architecture for the synthetic data set, the MNIST, fashion MNIST, and CelebA data sets. The architecture of the encoder is as follows,
\begin{enumerate}
    \item Conv Layer (32), Relu
    \item Conv Layer (64)
    \item Flatten (get flatten dimension F)
    \item Linear Layer (F to L)
\end{enumerate}
Where the numbers in the parenthesis determine the number of output channels. The decoder includes the following steps,
\begin{enumerate}
    \item Linear Layer (L to F2), (with $F_2=300\times D^0_i\times D^1_i$), and $D^0_i$, $D^1_i$ computed using the Conv. Transpose formulas on the following layers.
    \item Reshape (32, $D^0_i$, $D^1_i$)
    \item ConvT Layer (256), Relu
    \item ConvT Layer (128), Relu
    \item ConvT Layer (32), Relu
    \item ConvT Layer (8)
    \item Conv Layer (output channels, 1 or 3)
\end{enumerate}
Note that since we know the shape of the output and the design of the decoder, by using the formulas for how the kernel affects the image size on a convolution transpose, one can find $D^0_i$, and $D^1_i$, the closet height and width of the image to be given to the decoder to get the required output shape. Since the stride is not equal to one and the equations include the inverse of a floor operator, usually the exact output shape can not be found. Accordingly, we use the last Conv. Layer to fix both the number of channels and the image size to be compatible with the output. All Conv and ConvT layer except the last one have a kernel of size 3, stride of 2, padding of 1, and dilation of 1. The transpose convolutions also have an output padding of 1. The last Conv layer is simply to adjust the right number of channels and the right image size, so it has a stride of 1, and the kernel size is determined from the output of last ConvT layer as follows,
\begin{equation}
    k_{last\ conv} = 1 + (D^0_f-H)+(D^1_f-W),
\end{equation}
with $D^0_f$, and $D^1_f$ the dimensions of the image after the last Conv transpose, and $H$, $W$ are the height and width of the original image. Further, note that aRRAEs converged to a bottleneck of size $16$, $23$, and $186$ for the MNIST, fashion MNIST, and CelebA respectively. Accordingly, these are the bottleneck sizes used for RRAEs and Diabolo AEs for each of those problems. The training process was specified by the number of forward passes, the batch size, and the learning rate. For all of the problems presented in the table, the training parameters can be found below,

\begin{table}[!h]
  \centering
  \begin{tabular}{ccccc}
      \toprule
      Parameter         & Synthetic & MNIST                      & Fashion-MNIST                     & CelebA   \\
      \midrule
      Forward passes (Fine-tuning) & 2500 (500)                   & 20000  (500)                &  20000 (500)  & 500 (10)\\
      Learning Rate & $1\mathrm{e}{-3}$                   & $1\mathrm{e}{-4}$                & $1\mathrm{e}{-3}$ & $1\mathrm{e}{-3}$ \\
      Batch size     &  64          &
      32      & 48  &  4608 \\
      \bottomrule
  \end{tabular}
  \caption{Training parameters for the problems presented in the paper.}
  \label{training_param}
\end{table}
Note that the fine-tuning forward passes show for how many passes the basis for RRAEs were fixed and the decoder fine tuned. Note that the numbers in parenthesis do not mean additional iterations. For example, $2500 (500)$ means the RRAE was trained for $2000$ forward passes with an SVD and the encoder was then fixed for the last $500$ forward passes. Consequently, RRAEs were trained for the same number of forward passes as other models, but with their encoder fixed in the last few forward passes. Finally, note the choice of the batch size which is always a multiple of $8$. This is the case since all the trainings done for this paper where done on Google TPUs, in parallel over 8 TPUs. Hence, we decided to train CelebA on a large batch size for training to be more efficient, but we had to keep the batch size as a multiple of $8$ to be able to split every batch over 8 TPUs.

\section{Stagnation Criteria for aRRAEs}\label{aRRAEs details}
    In the following algorithm, we present the details of the stagnation criteria used for aRRAEs for all the  examples in the paper. Note that these are only a proposition, and any other suitable stagnation criteria can be used in practice.
    \begin{algorithm}[H]
  \caption{Stagnation Critirea for aRRAEs, with ``Forward Steps'' meaning number of batches doing ofrward passes (not epochs), $loss_s$ being the loss at Forward Step $s$, and $avg_s$ being the average of the losses over the last 20 forward passes.}
  \begin{algorithmic}
  \State stagnated = False,    steps = 0.
  
    \For{$s$ in Forward Steps}
    \If{$|loss_s-avg_s|/|avg_s|\times100< \epsilon_0$}
        \State steps $+= 1$
      \If{steps $> 500$}
        \State stagnated = True
      \EndIf
    \EndIf
    \EndFor
  \end{algorithmic}
\end{algorithm}


\newpage
\vskip 0.2in
\bibliography{sample}

\begin{thebibliography}{23}
\providecommand{\natexlab}[1]{#1}
\providecommand{\url}[1]{\texttt{#1}}
\expandafter\ifx\csname urlstyle\endcsname\relax
  \providecommand{\doi}[1]{doi: #1}\else
  \providecommand{\doi}{doi: \begingroup \urlstyle{rm}\Url}\fi

\bibitem[Abdi and Williams(2010)]{abdi2010principal}
Herv{\'e} Abdi and Lynne~J Williams.
\newblock Principal component analysis.
\newblock \emph{Wiley interdisciplinary reviews: computational statistics}, 2\penalty0 (4):\penalty0 433--459, 2010.

\bibitem[Alloghani et~al.(2020)Alloghani, Al-Jumeily, Mustafina, Hussain, and Aljaaf]{unsup_rev}
Mohamed Alloghani, Dhiya Al-Jumeily, Jamila Mustafina, Abir Hussain, and Ahmed~J Aljaaf.
\newblock A systematic review on supervised and unsupervised machine learning algorithms for data science.
\newblock \emph{Supervised and unsupervised learning for data science}, pages 3--21, 2020.

\bibitem[Arpit et~al.(2016)Arpit, Zhou, Ngo, and Govindaraju]{regularized_1}
Devansh Arpit, Yingbo Zhou, Hung Ngo, and Venu Govindaraju.
\newblock Why regularized auto-encoders learn sparse representation?
\newblock In \emph{International Conference on Machine Learning}, pages 136--144. PMLR, 2016.

\bibitem[Bank et~al.(2021)Bank, Koenigstein, and Giryes]{bank2021autoencoders}
Dor Bank, Noam Koenigstein, and Raja Giryes.
\newblock Autoencoders, 2021.
\newblock URL \url{https://arxiv.org/abs/2003.05991}.

\bibitem[Chandar et~al.(2014)Chandar, Lauly, Larochelle, Khapra, Ravindran, Raykar, and Saha]{speech_rec}
Sarath Chandar, Stanislas Lauly, Hugo Larochelle, Mitesh~M. Khapra, Balaraman Ravindran, Vikas Raykar, and Amrita Saha.
\newblock An autoencoder approach to learning bilingual word representations, 2014.
\newblock URL \url{https://arxiv.org/abs/1402.1454}.

\bibitem[Chen and Liu(2011)]{locally}
Jing Chen and Yang Liu.
\newblock Locally linear embedding: a survey.
\newblock \emph{Artificial Intelligence Review}, 36:\penalty0 29--48, 2011.

\bibitem[Chen and Guo(2023)]{AE_rev}
Shuangshuang Chen and Wei Guo.
\newblock Auto-encoders in deep learning—a review with new perspectives.
\newblock \emph{Mathematics}, 11\penalty0 (8):\penalty0 1777, 2023.

\bibitem[Chow and Liu(1968)]{chow:68}
C.~K. Chow and C.~N. Liu.
\newblock Approximating discrete probability distributions with dependence trees.
\newblock \emph{IEEE Transactions on Information Theory}, IT-14\penalty0 (3):\penalty0 462--467, 1968.

\bibitem[Deng(2012)]{deng2012mnist}
Li~Deng.
\newblock The mnist database of handwritten digit images for machine learning research.
\newblock \emph{IEEE Signal Processing Magazine}, 29\penalty0 (6):\penalty0 141--142, 2012.

\bibitem[JAX(2024)]{jax_jvp}
JAX.
\newblock The autodiff cookbook.
\newblock \url{https://docs.jax.dev/en/latest/notebooks/autodiff_cookbook.html}, 2024.

\bibitem[Jing et~al.(2020)Jing, Zbontar, et~al.]{implicit}
Li~Jing, Jure Zbontar, et~al.
\newblock Implicit rank-minimizing autoencoder.
\newblock \emph{Advances in Neural Information Processing Systems}, 33:\penalty0 14736--14746, 2020.

\bibitem[Kumar et~al.(2020)Kumar, Poole, and Murphy]{regularized_3}
Abhishek Kumar, Ben Poole, and Kevin Murphy.
\newblock Regularized autoencoders via relaxed injective probability flow.
\newblock In \emph{International conference on artificial intelligence and statistics}, pages 4292--4301. PMLR, 2020.

\bibitem[Lay(2012)]{lay2012linear}
David~C. Lay.
\newblock \emph{Linear Algebra and Its Applications}.
\newblock Pearson Education, Boston, 4th edition, 2012.
\newblock ISBN 978-0321385178.

\bibitem[Liu et~al.(2015)Liu, Luo, Wang, and Tang]{liu2015faceattributes}
Ziwei Liu, Ping Luo, Xiaogang Wang, and Xiaoou Tang.
\newblock Deep learning face attributes in the wild.
\newblock In \emph{Proceedings of International Conference on Computer Vision (ICCV)}, December 2015.

\bibitem[Nakatsukasa(2021)]{yuji}
Yuji Nakatsukasa.
\newblock C6.1 numerical linear algebra.
\newblock \url{https://courses.maths.ox.ac.uk/pluginfile.php/11960/mod_resource/content/17/NLA_lecture_notes.pdf}, December 2021.

\bibitem[Ng et~al.(2011)]{ng2011sparse}
Andrew Ng et~al.
\newblock Sparse autoencoder.
\newblock \emph{CS294A Lecture notes}, 72\penalty0 (2011):\penalty0 1--19, 2011.

\bibitem[Park et~al.(2018)Park, Hoshi, and Kemp]{anomaly}
Daehyung Park, Yuuna Hoshi, and Charles~C. Kemp.
\newblock A multimodal anomaly detector for robot-assisted feeding using an lstm-based variational autoencoder.
\newblock \emph{IEEE Robotics and Automation Letters}, 3\penalty0 (3):\penalty0 1544--1551, 2018.
\newblock \doi{10.1109/LRA.2018.2801475}.

\bibitem[Rifai et~al.(2011)Rifai, Vincent, Muller, Glorot, and Bengio]{CAE}
Salah Rifai, Pascal Vincent, Xavier Muller, Xavier Glorot, and Yoshua Bengio.
\newblock Contractive auto-encoders: explicit invariance during feature extraction.
\newblock In \emph{Proceedings of the 28th International Conference on International Conference on Machine Learning}, ICML'11, page 833–840, Madison, WI, USA, 2011. Omnipress.
\newblock ISBN 9781450306195.

\bibitem[Sch{\"o}lkopf et~al.(1997)Sch{\"o}lkopf, Smola, and M{\"u}ller]{scholkopf1997kernel}
Bernhard Sch{\"o}lkopf, Alexander Smola, and Klaus-Robert M{\"u}ller.
\newblock Kernel principal component analysis.
\newblock In \emph{International conference on artificial neural networks}, pages 583--588. Springer, 1997.

\bibitem[Vincent et~al.(2010)Vincent, Larochelle, Lajoie, Bengio, Manzagol, and Bottou]{SAE}
Pascal Vincent, Hugo Larochelle, Isabelle Lajoie, Yoshua Bengio, Pierre-Antoine Manzagol, and L{\'e}on Bottou.
\newblock Stacked denoising autoencoders: Learning useful representations in a deep network with a local denoising criterion.
\newblock \emph{Journal of machine learning research}, 11\penalty0 (12), 2010.

\bibitem[Wang and Tong(2022)]{IAE}
Xinyi Wang and Lang Tong.
\newblock Innovations autoencoder and its application in one-class anomalous sequence detection.
\newblock \emph{Journal of Machine Learning Research}, 23\penalty0 (49):\penalty0 1--27, 2022.

\bibitem[Xiao et~al.(2017)Xiao, Rasul, and Vollgraf]{xiao2017fashionmnistnovelimagedataset}
Han Xiao, Kashif Rasul, and Roland Vollgraf.
\newblock Fashion-mnist: a novel image dataset for benchmarking machine learning algorithms, 2017.
\newblock URL \url{https://arxiv.org/abs/1708.07747}.

\bibitem[Zhao et~al.(2018)Zhao, Kim, Zhang, Rush, and LeCun]{regularized_2}
Junbo Zhao, Yoon Kim, Kelly Zhang, Alexander Rush, and Yann LeCun.
\newblock Adversarially regularized autoencoders.
\newblock In \emph{International conference on machine learning}, pages 5902--5911. PMLR, 2018.

\end{thebibliography}

\end{document}